%% file: main.tex
\PassOptionsToPackage{table}{xcolor}
\documentclass[onecolumn]{openhelix}

\input{math_commands.tex}

\usepackage{amsmath}
\usepackage{amsfonts}
\usepackage{amssymb}
\usepackage{amsthm}
\usepackage{mathtools}
\usepackage{fdsymbol}
\usepackage{array}
\usepackage{tabularx}
\usepackage{longtable}
\usepackage{makecell}
\usepackage{xspace}
\usepackage{afterpage}
\usepackage{pifont}

\usepackage{wrapfig}
\usepackage{float}

\newcommand{\method}{UniWAM}

\title{UniWAM: Unified World-Action Model}
\author{\protect\makebox[\linewidth][c]{UniWAM Team}}

\hypersetup{
  pdftitle={UniWAM: Unified World-Action Model},
  pdfauthor={UniWAM Team}
}

\abstract{%
Vision-language-action models benefit from the understanding and reasoning capabilities of pretrained vision-language models, but action-only supervision provides limited grounding in world dynamics. Conversely, world-action models inherit spatiotemporal priors from video generation models, yet remain limited in semantic understanding and reasoning under distribution shifts. We introduce UniWAM, a unified architecture that integrates a physical reasoner, a world generator, and an action predictor to jointly learn semantic understanding of the physical world, visual generation, and action prediction. 
To ensure the quality of the training data, we developed a rigorous data cleaning and annotation pipeline for both human egocentric data and robot data.
To adapt the vision-language component to embodied tasks while preserving its inherited language capabilities, we represent low-level actions in natural language and introduce a pre-training recipe that assigns complementary supervision from visual question answering (VQA) data, human egocentric data, and robot demonstrations to the appropriate model components. 
During post-training, future visual noise augmentation reduces reliance on precise future predictions, while history-conditioned flow matching uses encoded action history to initialize action generation. Together, these designs significantly reduce denoising steps while maintaining performance.
UniWAM achieves state-of-the-art (SOTA) performance across multiple evaluations, including in-distribution performance, robustness, generalization, instruction following, and long-horizon task execution.
Furthermore, we uncover a log-linear scaling law of unified human-robot co-training, demonstrating the effectiveness of large-scale pre-training on a mixture of human and robot data.
}

\newcommand{\githubicon}{\raisebox{-1.5pt}{\includegraphics[height=1.05em]{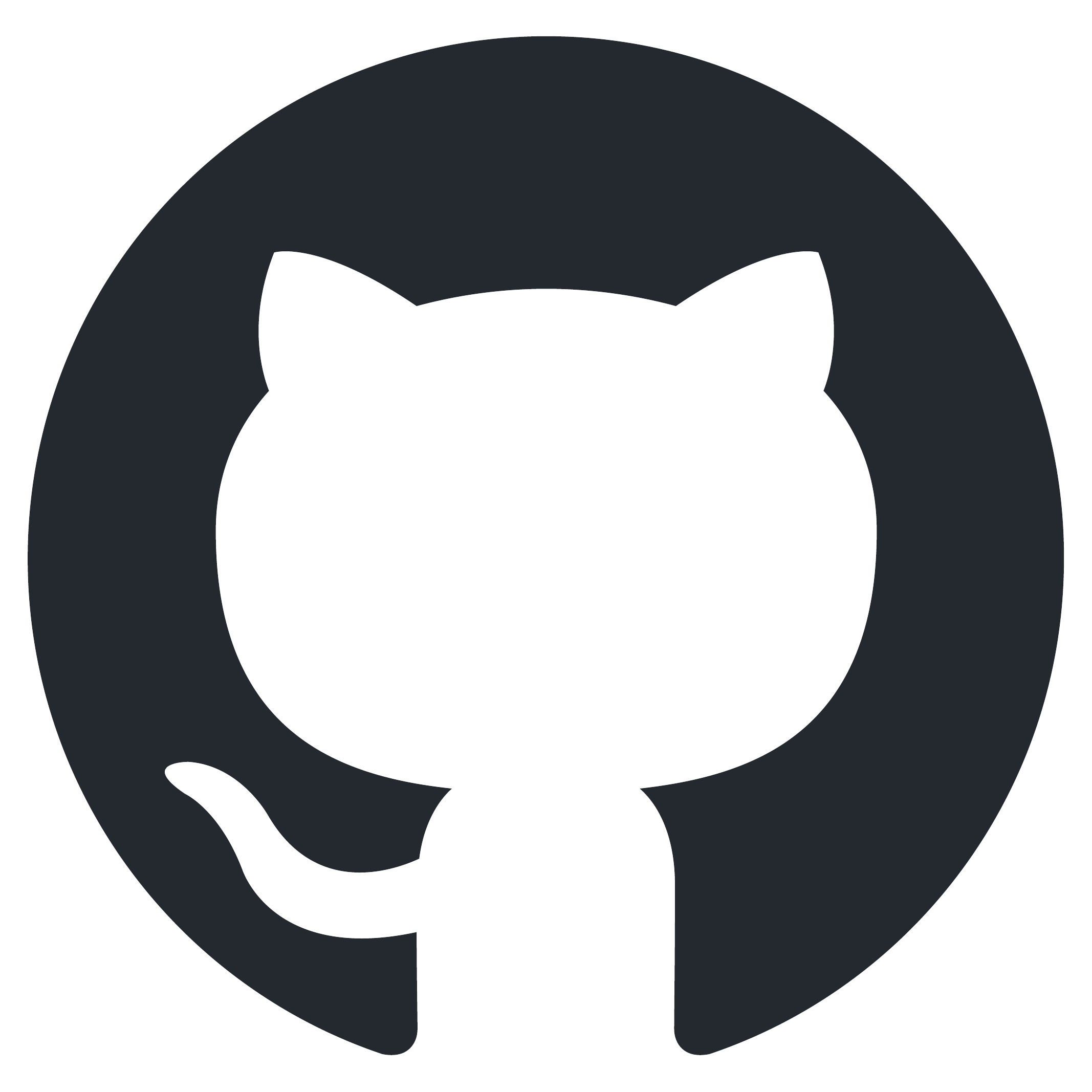}}\xspace}
\newcommand{\huggingfaceicon}{\raisebox{-1.5pt}{\includegraphics[height=1.05em]{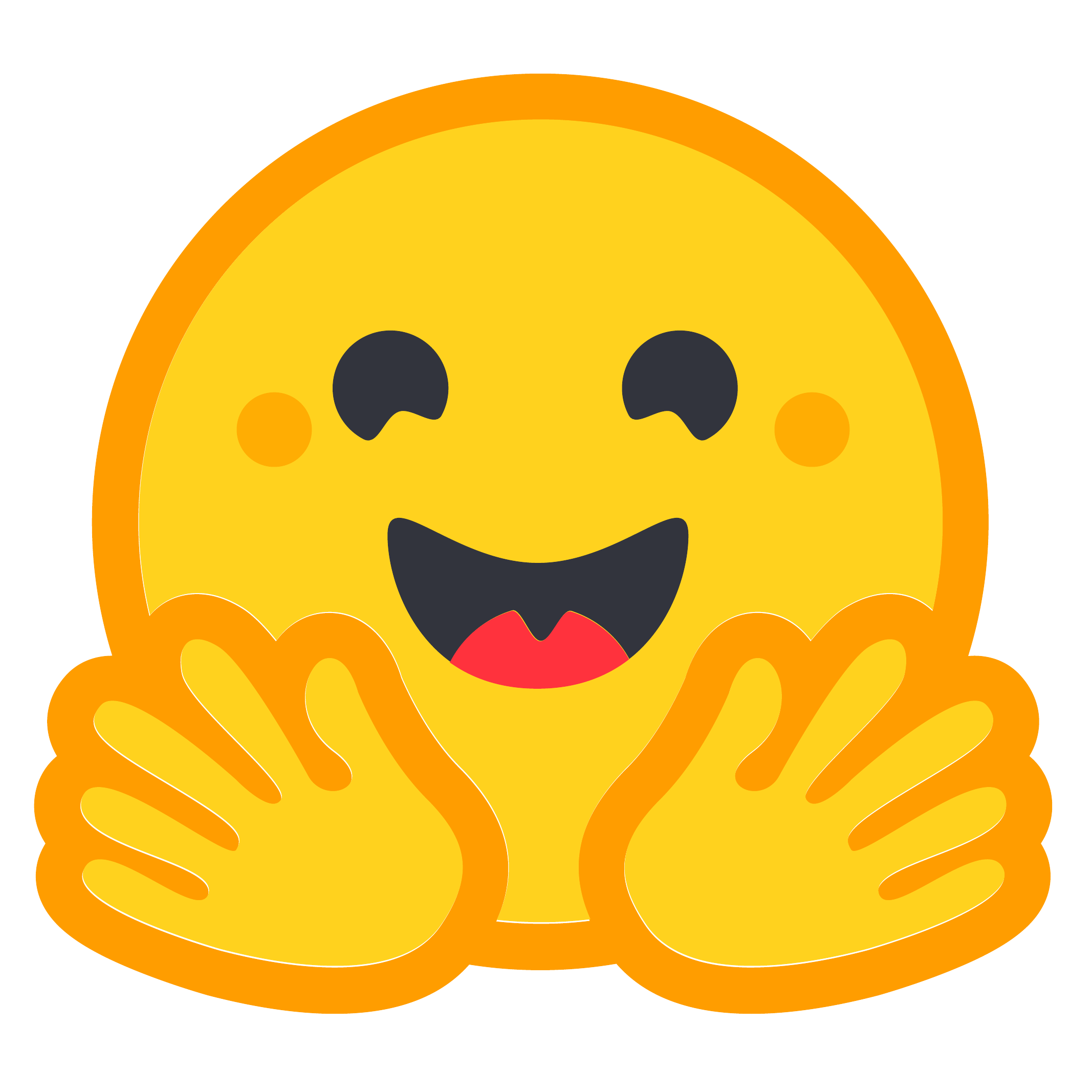}}\xspace}
\newcommand{\modelscopeicon}{\raisebox{-1.5pt}{\includegraphics[height=1.05em]{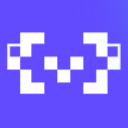}}\xspace}
\newcommand{\projecticon}{\raisebox{-1.5pt}{\includegraphics[height=1.05em]{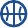}}\xspace}

\metadata[Code]{\href{https://github.com/UniWAM/UniWAM}{\githubicon~\textcolor[HTML]{1F4E8C}{GitHub}}}
\metadata[Model]{\href{https://huggingface.co/collections/chenpyyy/uniwam}{\huggingfaceicon~\textcolor[HTML]{1F4E8C}{Hugging Face}}\quad\href{https://www.modelscope.cn/collections/Kosmos524/UniWAM}{\modelscopeicon~\textcolor[HTML]{1F4E8C}{ModelScope}}}
\metadata[Project Page]{\href{https://uniwam.github.io}{\projecticon~\textcolor[HTML]{1F4E8C}{Project Page}}}
\begin{document}

\maketitle

\begin{figure}[H]
    \centering
    \includegraphics[width=\textwidth]{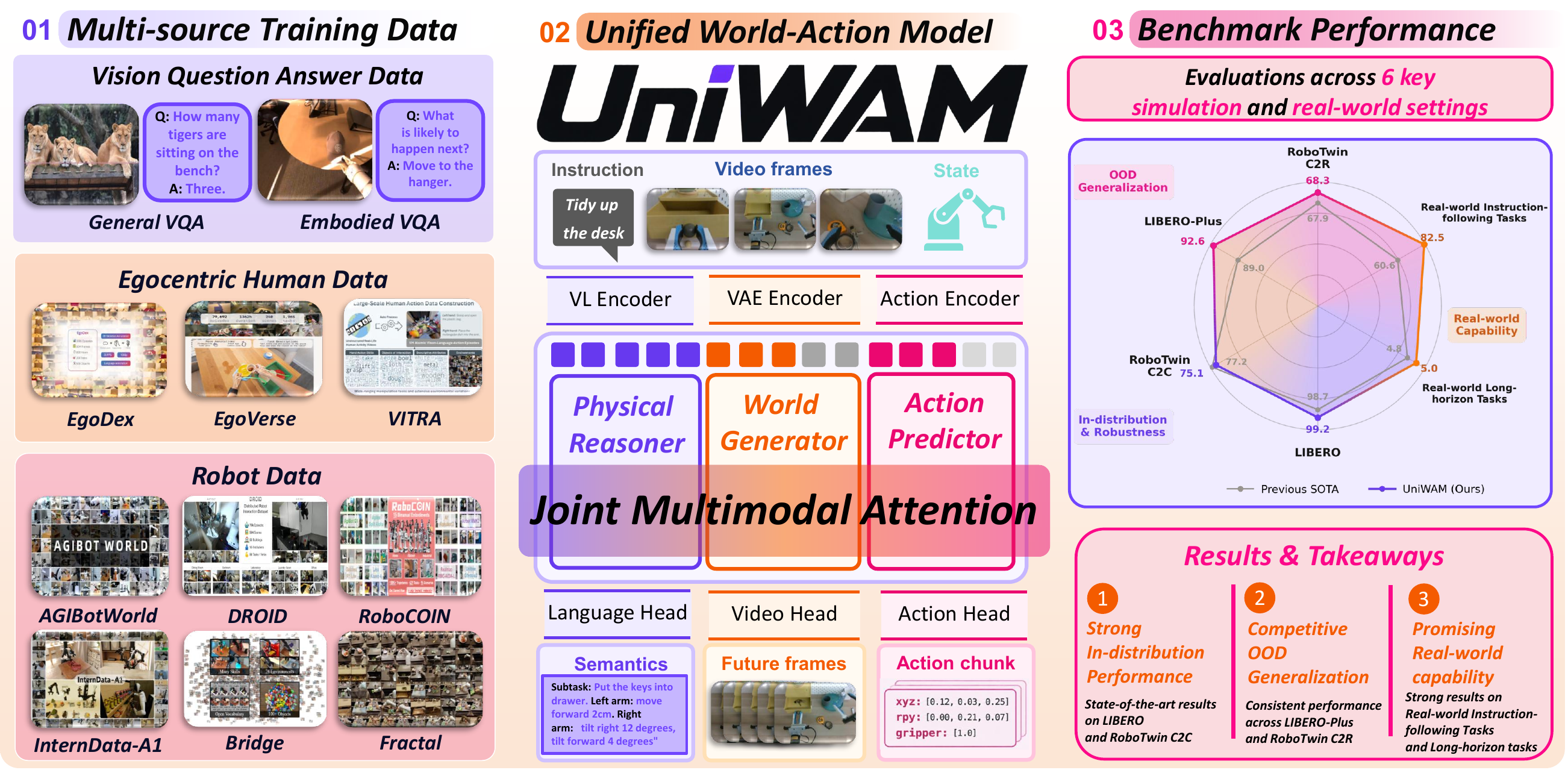}
    \caption{ 
    \textbf{\method}~unifies physical reasoning, visual generation, and action prediction in a unified architecture and pretrains on robot and human egocentric data as well as VQA data.
    The resulting model exhibits strong robustness, generalization, instruction-following, and reasoning abilities.
    }
    \label{fig:uniwam-teaser}
\end{figure}

\clearpage
\setcounter{tocdepth}{3}
\tableofcontents
\clearpage

\input{sec/01_introduction}
\input{sec/02_related_work}

\input{sec/03_method}
\input{sec/04_experiments}
% \input{sec/05_discussion}%只在tech report里出现，集思广益
\input{sec/06_conclusion}

\newpage

\section*{Authors}
\noindent\textbf{Core Contributors:} Jiayi Chen\textsuperscript{1,2}, Jingbo Wang\textsuperscript{1,2}, Shuai Zhou\textsuperscript{3}, Xicheng Gong\textsuperscript{4}

\noindent\textbf{Contributors:} Ziyang Zhou\textsuperscript{1,2}, Zehua Fan\textsuperscript{5}, Junwu E\textsuperscript{1}, Haodong Yan\textsuperscript{1}, Fuhao Li\textsuperscript{1,2}, Qize Yu\textsuperscript{4}, Xu Huang\textsuperscript{2}, Pengwei Wang\textsuperscript{6}, Wen Chen\textsuperscript{2}, Haoang Li\textsuperscript{1,2}

\noindent\textbf{Project Lead:} Wenxuan Song\textsuperscript{1}

\noindent\textbf{Project PI:} Shunbo Zhou\textsuperscript{2}

\medskip
{\small
\noindent\textsuperscript{1}The Hong Kong University of Science and Technology (Guangzhou)\par
\noindent\textsuperscript{2}OLA Dimensions\par
\noindent\textsuperscript{3}Carnegie Mellon University\par
\noindent\textsuperscript{4}Peking University\par
\noindent\textsuperscript{5}Shanghai Jiao Tong University\par
\noindent\textsuperscript{6}Beijing Academy of Artificial Intelligence\par
% \noindent\textsuperscript{8}The Chinese University of Hong Kong\par

}

% \subsubsection*{AI use statement}
% Generative AI tools were used to polish the manuscript, organize its structure,
% and assist with code writing. They were not used to generate experimental data,
% execute the reported evaluations, or generate references. We reviewed all AI-assisted work and take responsibility
% for the final content of this paper, including any text, claims, or artifacts
% produced with the aid of generative AI.

% \subsubsection*{Ethics statement}
% This work studies simulated and planned real-world robotic manipulation. The
% current experiments do not involve human participants or personal data. A
% deployed system could nevertheless cause physical damage when instructions are
% misinterpreted or control fails under distribution shift. Any real-world
% evaluation should therefore use workspace constraints, emergency-stop
% procedures, and human supervision appropriate to the platform.

% \subsubsection*{Reproducibility statement}
% The manuscript records the model formulation and the configuration values that
% can be verified from the current implementation. Appendix~\ref{app:config}
% collects these values, and Appendix~\ref{app:task-results} reports all available
%  RoboTwin task-level results. The code and instructions for use are also included in the supplementary-materials archive.

\addcontentsline{toc}{section}{References}
\bibliographystyle{plainnat}
\bibliography{iclr2027_conference}

\newpage
\appendix
\section*{Appendix}
\addcontentsline{toc}{section}{Appendix}
\input{sec/appendix}

\end{document}

%% file: math_commands.tex
\usepackage{amsmath,amsfonts,bm}

\def\eqref#1{equation~\ref{#1}}
\def\1{\bm{1}}

\DeclareMathAlphabet{\mathsfit}{\encodingdefault}{\sfdefault}{m}{sl}
\SetMathAlphabet{\mathsfit}{bold}{\encodingdefault}{\sfdefault}{bx}{n}

%% file: sec/01_introduction.tex
\section{Introduction}
\label{sec:introduction}

% 近年来，以VLM作为backbone构建出的机器人基础模型，即VLA很有效，其能有效利用VLM强大预训练带来的多模态理解能力，但是单纯以动作作为监督目标导致模型对dynamics的理解能力差。WAM inherit strong spatiotemporal priors from video-generation backbones pre-trained on large-scale
% video data, 通过predict next frame的方式提升了dynamics，但是其OOD场景下的understanding以及complex任务下的reasoning能力差。This paper asks: 我们如何将VLM强大的推理和泛化能力与WAM强大的dynamics理解能力相结合，从而构建真正的robot
% generalist?
In recent years, robot foundation models built upon Vision-Language Models (VLMs)~\citep{bai2025qwen3vl, beyer2024paligemma,
an2025llava}, namely Vision-Language-Action Models (VLAs)~\citep{physicalintelligence2025pi05, generalist2025gen0, bai2026embodied}, have
demonstrated strong effectiveness, as they can effectively leverage the semantic understanding capabilities
acquired through large-scale VLM pre-training. However, using actions
alone as the supervision signal limits the model's ability to
understand world dynamics. World-Action Models (WAMs)~\citep{ye2026dreamzero, bi2025motus, li2026lingbotva, yang2026four_d_wam} inherit strong spatiotemporal
priors from Video-Generation Models (VGMs)~\citep{wan2025wan, li2026xiaomi} pre-trained on
large-scale video data, improving their understanding of dynamics through predicting the next
frame, but their understanding in out-of-distribution (OOD) scenarios
and reasoning capabilities for complex tasks remain limited. This paper
asks: How can we combine the strong reasoning and generalization capabilities of
VLMs with the strong dynamics understanding of WAMs,
thereby building a truly generalist robot?

% To address this question, we introduce UniWAM, 桥接强大的VLM和VGM在一个统一的Unified 架构内,借助了MoT的思想~\cite{MoT, Motus}. 预训练这种混合架构面临两个关键问题：1.
% 为了让VLM适应具身任务~\cite{vla-adapter}，我们希望vlm不是冻结的，同时为了保持其强大的语言能力，我们需要给vlm合适的监督信号. 为此，我们展示直接使用自然语言表示低层机器人动作，使VLM的动作监督信号与预训练的输入输出分布保持一致，是有效的训练VLM理解能力并保持预训练能力的一种方式 2. 为了利用尽可能多样的数据，同时保持vlm的语言能力，我们需要构建多样的数据源，并且构建合理的co-training recipe。为此，我们构造了包含（vqa， robot， human）的预训练数据源，并合理的分配三者参与监督的方式。即VQA用于监督VLM, Human
% Data用于监督VLM和VGM，从而充分利用其提供的物理规律学习、并避免收到低精度的action label的标注，由于Robot Data和后训练的物理一致性最高，动作精度最高，因此Robot Data用于监督VLM, VGM和Action Expert，训练精细的action output能力。
To address this question, we introduce UniWAM, integrating a VLM-based physical reasoner,
a VGM-based world generator, and an action predictor
within a unified architecture, drawing inspiration from the idea of Mixture-of-Transformer (MoT)~\citep{liang2025mot,bi2025motus}.
% Ths hybird model 同时具备Semantic Understanding, Visual Prediction, and Action Generation的能力。
The hybrid model jointly supports semantic understanding, visual prediction, and action generation.
Figure~\ref{fig:uniwam-teaser} summarizes the training data, unified architecture,
and benchmark performance of \method.
\textbf{Pre-training} such a hybrid model presents three key challenges:
1) Adapting the VLM to embodied tasks~\citep{wang2026vlaadapter} requires updating its parameters to
acquire embodied knowledge. However, such adaptation must preserve
the language capabilities inherited from pretraining, which calls for supervision signals that
remain compatible with the VLM's pretraining distribution. To
this end, we show that directly representing low-level robot actions
using natural language~\citep{zha2026lap}, thereby aligning the VLM's action supervision with its pretraining
input-output distribution, provides an effective way to train
the VLM for embodied understanding while preserving its pretrained capabilities.
% 2) To leverage data from as diverse sources as possible while
% preserving the VLM's language capabilities, we need to
% construct diverse data sources and develop an appropriate co-training recipe.
2) Another key challenge lies in designing appropriate supervision signals for each
expert using heterogeneous data, as differences in data
distributions and training objectives may lead to conflicts among the optimization processes
of different experts.
To this end, we construct a pretraining mixture consisting of Visual Question Answering (VQA), robot data, and human data, and
carefully assign their supervision to different components. Specifically, VQA data supervises the VLM to maintain its pretraining knowledge. Human data supervises both the VLM and VGM, thereby fully exploiting the physical knowledge contained in human videos while avoiding supervision from low-precision action
labels. 
Since robot data exhibits the highest physical
consistency with downstream post-training and provides the most accurate action annotations, it
is used to supervise the VLM, VGM, and
action predictor, enabling the learning of precise action outputs.
3) 
%大规模异构数据混合训练对数据质量提出了极高的要求。因此我们developed a rigorous data cleaning pipeline。对于人类数据来说，一条轨迹往往包含若干个动作，因此我们进一步为人类数据设计了自动的轨迹切分和标注管线。
Large-scale training on heterogeneous data places stringent demands on data quality. We therefore developed a rigorous data cleaning pipeline. Since individual trajectories in human data often contain multiple actions, we further designed an automated pipeline for segmenting and annotating human trajectories.

% 在后训练中我们同时监督三个以延续预训练共同unified加强模型理解生成预测三个的思想，
% 同时我们引入了未来视觉噪声增强和历史条件流匹配。前者在保留条件帧的同时，对未来视觉潜变量施加部分噪声扰动，使动作分支学习从粗略的视觉表示中提取与控制相关的语义信息，降低对精确未来视觉预测的依赖。
% 进一步的，动作分支利用历史序列中的动力学信息与时间连续性将历史动作序列映射到高维潜在空间，并以此作为流匹配的生成起点，使历史运动信息引导未来动作轨迹的生成，保持性能的同时将去噪步数从 10 步减少至2步，大幅提高了推理效率。
During \textbf{post-training}, we continue the pre-training principle of maintaining unified capabilities and
jointly supervise the three experts
to adapt the model's understanding, generation, and prediction capabilities to specific embodied
scenarios.
We additionally introduce future-visual noise augmentation and history-conditioned flow matching.
% The former partially perturbs future visual latents with noise, encouraging the
% action branch to extract control-relevant semantic information from
% coarse visual representations and reducing its reliance on precise future visual
% predictions.
The former partially perturbs future visual latents while preserving the conditioning frames,
encouraging the action expert to extract control-relevant semantics
from coarse visual representations rather than rely on precise future predictions, thus
strengthening the interaction between visual generation and action
prediction.
Furthermore, the action expert leverages the dynamics and temporal continuity encoded in
historical sequences. It maps the action history into
a high-dimensional latent space and uses the resulting representation to initialize the
flow-matching process, thereby allowing historical motion information to
guide the generation of future action trajectories.
Together, these designs substantially reduce the number of denoising steps while maintaining
performance, improving inference efficiency.
Overall, at the architectural level, we build a MoT that connects language
reasoning, video generation, and action prediction through joint
attention.
Our contributions are summarized in four aspects.
\begin{itemize}
    % \item At the architectural level, we establish a unified training objective
    % for semantic understanding, visual prediction, and action generation.
    \item For pre-training, we construct a three-source pre-training scheme comprising human data,
    robot teleoperation data, and VQA data, and systematically
    evaluate the contribution of each data source through source-level ablation studies.
    \item For post-training, we explicitly formulate how recent action history and perturbed
    future visual latents are incorporated into the joint
    flow-matching process.
    \item Extensive evaluations demonstrate that our \method~achieves SOTA performance across multiple dimensions,
    including in-distribution performance, robustness, generalization, instruction following, and
    long-horizon task execution. 
    \item We uncover a log-linear scaling law of unified human-robot co-training, demonstrating the effectiveness of large-scale pre-training on a mixture of human and robot data.
    % Extensive evaluations demonstrate that UniWAM achieves the highest overall success rate
    % among the compared methods on the standard LIBERO
    % benchmark. On the complete 50-task RoboTwin 2.0 benchmark, UniWAM achieves average
    % success rates of 75.14\% and 68.32\% under the
    % clean2clean and clean2rand settings, respectively~\citep{chen2025robotwin2}. Across the four standard LIBERO suites,
    % UniWAM
    % successfully completes 1,980 out of 2,000 evaluation episodes, yielding an overall
    % success rate of 99.0\%. Under the seven perturbation
    % categories of LIBERO-Plus, UniWAM achieves an aggregate success rate of 92.58\%.
\end{itemize}

%% file: sec/02_related_work.tex
\section{Related Work}
\label{sec:related-work}

\noindent\textbf{Large-scale Pretraining for Robotics.}
Large-scale robotic pretraining has brought the field closer to generalist robots capable of performing a broad range of tasks across diverse environments.
RT-2~\citep{brohan2023rt2} jointly trains on web
and robot data with tokenized actions, while
OpenVLA~\citep{kim2024openvla} adapts pretrained
vision--language representations to robot
demonstrations. $\pi_0$~\citep{black2024pi0}
first combines a vision--language backbone with flow
matching.
$\pi_{0.5}$~\citep{physicalintelligence2025pi05}
adds semantic subtask supervision and web data.
GEN-0~\citep{generalist2025gen0} leverages Unified Manipulation Interface (UMI) data to acquire physical sense towards scaling laws, while
Xiaomi-Robotics-1~\citep{xiaomi2026robotics1}
further constructs a two-state pretraining on UMI data for physical understanding and teleoperation data for embodiment-specific adaptation. 
Galaxea G0.5~\citep{liu2026g05} unifies reasoning and
action tokens through autoregressive training on
robot and visual question answering data.
Motus~\citep{bi2025motus} learns shared motion
information from heterogeneous data using
optical-flow-derived latent actions. 
However, these approaches primarily emphasize either language–action or video–action joint modeling.
In contrast, our \method's pretraining framework unifies semantic understanding, dynamics prediction, and action generation through diverse data and complementary supervision.

\noindent\textbf{World-action Models.}
World-action models (WAMs) are typically built
upon video generation backbones. They leverage the
physical dynamics priors of large-scale video
pretraining and obtain dense supervision by
predicting the future in diverse representation
spaces---including pixels, geometry, and
value---thereby facilitating the learning of
stronger action-generation policies.
DreamZero~\citep{ye2026dreamzero}, Cosmos
Policy~\citep{kim2026cosmospolicy},
Motus~\citep{bi2025motus},
MotuBrain~\citep{xiang2026motubrain},
LingBot-VA~\citep{li2026lingbotva}, and
Dyna-2~\citep{dyna2026dyna2} focus on modeling
visually grounded world dynamics and aligning
visual-state evolution with actions.
Fast-WAM~\citep{yuan2026fastwam},
GigaWorld-Policy~\citep{ye2026gigaworldpolicy},
ImageWAM~\citep{zhang2026imagewam}, and
S-VAM~\citep{yan2026svam} further show that the
video-generation backbone can serve at inference
time as a representation encoder only, without
synthesizing complete future frames.
4D-WAM~\citep{yang2026four_d_wam} and
X-WAM~\citep{guo2026xwam} extend the
representation domain of WAMs from pixel-temporal
features to four-dimensional spatiotemporal
geometry. MobileWAM~\citep{fan2026mobilewam} and
ABot-M0.5~\citep{chen2026abotm05} further push
WAMs toward more complex mobile manipulation.
Unlike these works, which primarily seek to unlock
the capabilities of video generation models, we
unify a vision--language model, a video generation
model, and an action expert within a single
framework and optimize it with complementary data
and supervision, enabling \method{} to jointly
support multimodal understanding, visual
generation, and action prediction.

\noindent\textbf{Learning from Egocentric Data.}
Egocentric videos provide a scalable source of
embodied experience beyond costly robot
demonstrations. Recorded from the actor's
viewpoint, they capture diverse human--object
interactions and the resulting changes in the
physical world, while covering a broad range of
objects, environments, and long-horizon
activities. Recent datasets further enrich raw
videos with language, human motion, geometry, and
action-related signals, making them increasingly
suitable for large-scale embodied
pretraining~\citep{ego4d,epickitchen,egodex,%
hoi4d,egoverse,openaoe,egolive,humannet}. On the
modeling side, early efforts mainly focused on
representation learning, while later work
introduced more structured supervision through
latent actions, future prediction, and geometric
motion
cues~\citep{lapa,gr00tn1,univla,jala,rynnvla,%
rotvla,flare,dreamdojo}. Moving closer to policy
learning, human motions and manipulation
trajectories are reconstructed and aligned with
robot action spaces for VLA
pretraining~\citep{egomimic,beingh0,hrdt,egovla,%
scalablehumanvla,metis,beingh05,egoscale}. Recent
embodied foundation models further unify human and
robot experience through shared action
representations and cross-embodiment pretraining,
allowing egocentric data to support both scalable
action learning and world
modeling~\citep{beingh05,egoscale,aceego,%
lingbotvla2,egosteer,ego2robot,riemann1}.

%% file: sec/03_method.tex
\section{Data}
\input{table/dataset_durations.tex}

\subsection{Data Source}

\subsubsection{Robot Data}
Our robot pretraining data combine
AgiBotWorld2026~\citep{agibotworld2026},
Bridge~\citep{walke2023bridgedata},
Droid~\citep{khazatsky2024droid},
Fractal~\citep{brohan2023rt1},
Robocoin~\citep{robocoin2025}, and
Interdata-a1~\citep{interndata2025},
covering single-arm and dual-arm embodiments.
The selected data total approximately 4,363 hours, with a per-source
breakdown in Table~\ref{tab:dataset_durations}.
We also apply language-action supervision to these robot data,
using descriptions of local manipulation behavior.

\subsubsection{Human Egocentric Data}
% so we use egoverse egodex and vitra here and describe why we choose them and how do we wash them and potentially mention our annotation pipeline will be present in the appendix?
Our egocentric pretraining data are drawn from three large-scale sources.
EgoDex~\citep{egodex} captures dexterous tabletop manipulation through
egocentric videos paired with dense 3D hand and finger tracking.
EgoVerse~\citep{egoverse} comprises human demonstrations collected across
diverse real-world settings and participants, accompanied by language
and human-motion annotations.
VITRA~\citep{scalablehumanvla} transforms in-the-wild human activity videos
into atomic manipulation segments with language descriptions and
reconstructed hand and camera trajectories; we use its processed subsets
derived from Ego4D~\citep{ego4d}, Ego-Exo4D~\citep{grauman2024ego}, and
EPIC-KITCHENS~\citep{epickitchen}.
Together, these sources provide approximately 5,000 hours of egocentric
video spanning a broad range of manipulation tasks, objects, and
environments, with rich language and dense motion annotations. We use subtask-level annotations to align language descriptions with
the local manipulation behavior in each video segment, rather than
only the overall episode goal.
EgoVerse and VITRA already provide such annotations.
For EgoDex, we develop a VLM-based automatic annotation pipeline
that segments recordings into atomic manipulation subtasks and generates
a concise description for each subtask;

%@家毅，周帅，西诚
%已有研究表明，视觉语言联合训练有助于保留预训练 VLM 的理解能力并改善策略泛化，而面向具身推理的适配也能够为下游动作学习带来收益。基于此，我们在预训练中引入多源视觉问答和具身问答数据，旨在保留理解VLM的视觉语义知识，同时补充与机器人操作相关的感知和推理监督。在空间与场景理解方面，我们结合 SAT~\citep{sat}、RoboPoint~\citep{robopoint}、RefSpatial~\citep{roborefer}、VST-P~\citep{visual_spatial_tuning} 和 SenseNova-SI~\citep{sensenova_si}，覆盖物体指代、空间关系、距离与深度估计，以及跨视角对应等内容。此外，CLEVR~\citep{clevr} 和 GRiD-3D~\citep{grid3d}分别补充基于物体属性与关系的组合推理，以及物体自身参考系下的相对方向判断。

%为进一步覆盖时序变化与操作过程，我们纳入 VSI-590K~\citep{cambrian_s} 和 SIMS-VSI~\citep{sims_v}，提供视频中的空间与时序推理监督；同时结合 RoboVQA~\citep{robovqa} 和 Robo2VLM~\citep{robo2vlm}，补充任务状态判断和目标条件下的交互理解。在联合训练中，这些问答通过回答预测为理解专家提供语言监督，与来自人类和机器人数据的语言动作描述形成互补，并采用第~3.2.3~节所述的自回归语言目标进行优化。
\input{table/vqa_data}
\subsubsection{VQA Data}
We incorporate multi-source visual question answering into pretraining
to build visual--semantic and embodied reasoning capabilities for
downstream policy learning.
These question--answer pairs provide explicit language supervision
to help preserve the understanding expert's pretrained semantic
knowledge while strengthening its understanding of manipulation-relevant
scenes and interactions~\citep{lin2026systematic,gong2026extending,wan2026cometvla}.
For spatial and scene understanding, we combine SAT~\citep{ray2024sat},
RoboPoint~\citep{yuan2024robopoint}, RefSpatial~\citep{zhou2026roborefer},
VST-P~\citep{yang2025visual}, and SenseNova-SI~\citep{cai2026scaling},
covering object grounding, spatial relations, depth and distance
reasoning, and cross-view correspondence.
CLEVR~\citep{johnson2017clevr} and GRiD-3D~\citep{lee2022right}
further provide supervision for compositional reasoning over object
attributes and relations, and relative directions in object-intrinsic
reference frames, respectively.
To extend supervision to temporal changes and manipulation processes,
we include VSI-590K~\citep{yang2026cambrian}
and SIMS-VSI~\citep{brown2025sims}
for spatial and temporal reasoning in videos, together with
RoboVQA~\citep{sermanet2024robovqa}
and Robo2VLM~\citep{chen2025robo2vlm}
for task-state assessment and goal-conditioned interaction understanding.
During co-training, VQA responses supervise the understanding expert
through the autoregressive language objective described in
Section~3.2.3, complementing the language--action descriptions
derived from human and robot data.

\subsection{Data Processing}
Robot datasets collected from different platforms and acquisition
pipelines often exhibit heterogeneous failure modes in both trajectory
signals and visual observations.
We therefore apply a unified quality-control procedure before training.
Our pipeline evaluates each trajectory from three complementary
perspectives: temporal and statistical reliability, geometric
consistency, and visual validity.

\subsubsection{Robot Data Processing}

\noindent\textbf{Temporal and Statistical Trajectory Screening.}
\textit{(1) Abrupt-transition screening:}
For each scalar state or action trajectory, we construct a smooth reference using cascaded median filtering followed by Savitzky--Golay smoothing. A time step is flagged as anomalous when its residual from the reference exceeds a threshold and either its acceleration or jerk also exceeds the corresponding threshold. This joint criterion suppresses isolated non-physical spikes while remaining tolerant to gradual motion changes.
\textit{(2) State--action temporal consistency.}
For dimensions shared by the state and action representations, we smooth both trajectories and estimate their relative temporal offset using cross-correlation. After lag compensation, we compute directional agreement from their first-order differences and reject episodes below a dataset-specific threshold. For incremental action representations, actions are first integrated into absolute trajectories before comparison.
\textit{(3) Distribution-based outlier removal.}
We further remove rare numerical outliers using robust per-dimension statistics. Let $q_{01}$ and $q_{99}$ denote the $1$st and $99$th percentiles. Values are retained within
$[q_{01}-\alpha(q_{99}-q_{01}),q_{99}+\alpha(q_{99}-q_{01})]$.
Gripper dimensions are excluded because their distributions are typically discrete or bimodal.

\noindent\textbf{Geometry-Aware Consistency Checks.}
\textit{(1) Joint--EEF consistency:}
When reliable joint measurements, robot models, and end-effector semantics are available, we verify the consistency between recorded joint states and logged end-effector poses through forward kinematics. This check is used to detect discrepancies caused by joint-angle conventions, TCP definitions, rotation representations, or base-frame assumptions, and is skipped when the required metadata are ambiguous or incomplete.
\textit{(2) Coordinate-frame and orientation alignment:}
For datasets with explicitly defined coordinate semantics, we apply dataset-specific rigid-frame corrections to align observations to a common convention in which the positive $x$-axis corresponds to the robot's forward direction. When the original frame semantics are not sufficiently specified, the recorded orientation representation is retained unchanged.

\noindent\textbf{Visual Observation Quality.}
We remove visually invalid observations, including black, corrupted, severely blurred, and prolonged static frames. Static segments are identified jointly from visual, state, and action signals to distinguish redundant observations from task-relevant transitions. Interaction-critical frames, such as gripper-closure events, are preserved even when their visual change is small.

\subsubsection{Human Egocentric Data Processing and Annotation}
\label{sec:human-data-processing}
\label{sec:human-annotation}

We use subtask-level annotations to align language descriptions with
the local manipulation behavior in each video segment, rather than
only with the overall episode goal.
We retain the annotations provided by EgoVerse and VITRA and supplement
EgoDex with automatically generated subtask annotations.
The annotated videos are then processed into a common format with
temporally aligned visual observations, language descriptions, and
human motion trajectories.

\begin{figure}[t]
    \centering
    \includegraphics[width=\linewidth]{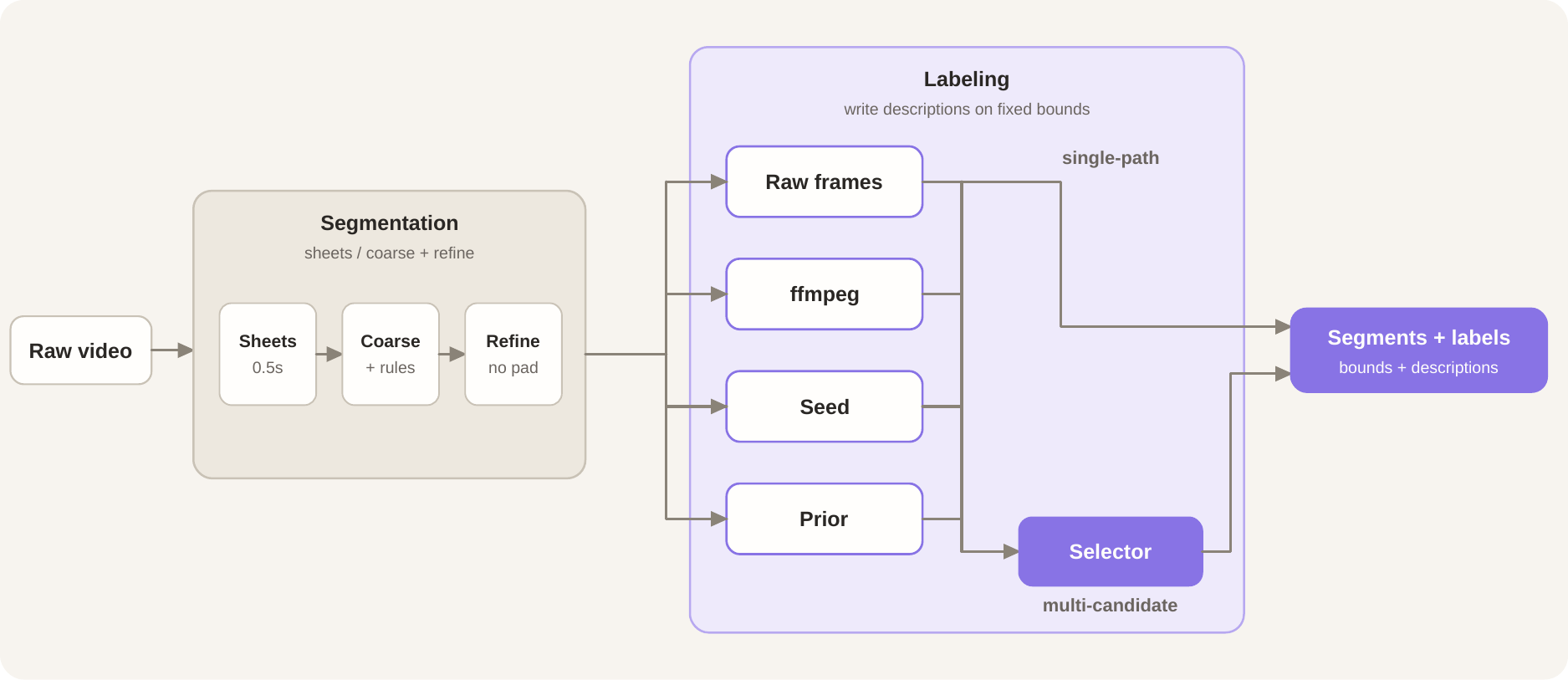}
    \caption{\textbf{EgoANT annotation pipeline.}
    An untrimmed egocentric video is represented as timestamped contact
    sheets and segmented through a full-episode coarse pass followed
    by local temporal refinement.
    With the refined boundaries fixed, the multi-candidate annotation
    route generates descriptions using raw-frame annotation,
    an alternative FFmpeg decoding and frame-sampling path, and
    relabeling conditioned on seed or prior descriptions.
    A text-only selector chooses the final description without
    re-examining the video.
    The output consists of temporal intervals paired with concise
    subtask descriptions, which we use to annotate EgoDex.}
    \label{fig:egoant-pipeline}
\end{figure}

\noindent\textbf{Automatic annotation of egocentric videos.}
We develop \textbf{EgoANT}, a VLM-based pipeline that converts long,
untrimmed egocentric videos into temporally localized manipulation
events and their language descriptions.
Given an episode $V$, the pipeline produces a sequence
$\{(t_i^{s}, t_i^{e}, y_i)\}_{i=1}^{N}$, where
$[t_i^{s}, t_i^{e}]$ denotes the temporal extent of an atomic
manipulation event and $y_i$ describes the completed action.
As illustrated in \Cref{fig:egoant-pipeline}, EgoANT separates this
process into two stages:
\emph{coarse-to-fine temporal segmentation}, which determines when
individual manipulation events occur, and \emph{semantic annotation},
which determines what action is performed in each resulting segment.

\noindent\textbf{Coarse-to-fine temporal segmentation.}
We sample one frame every $0.5$\,s and organize consecutive frames into
timestamped contact sheets with up to 20 frames per sheet.
This representation preserves the episode's temporal context while
providing explicit timestamps for boundary prediction.
A VLM processes the contact sheets for the full episode in a single
pass, together with the completed-event segmentation rules released
by Macrodata Labs~\citep{macrodata2026subtasks}.
These rules emphasize manipulation events that produce meaningful
object-state changes, such as picking, placing, opening, and pouring,
rather than treating approach motions, minor adjustments, or hand
withdrawals as independent subtasks.
The full-episode prediction provides coarse proposals rather than
final boundaries.
We subsequently construct local temporal windows around the coarse
boundaries and represent each window with timestamped contact sheets.
The VLM re-examines these local observations together with the coarse
hypothesis and episode-level context, without adding temporal padding
outside the selected window.
The refinement prompt asks the model to identify completed manipulation
events whose starts and ends are visible within the window; it does
not require the predicted segments to fill the entire window.
This second pass revises the temporal boundaries and can split an
under-segmented coarse proposal into multiple atomic subtasks.

\noindent\textbf{Segment-level semantic annotation.}
Once the temporal boundaries are fixed, we generate a concise
description of the completed manipulation in each segment, including
the manipulated object and its destination or resulting state when
observable.
Rather than relying on a single annotation pass, we construct multiple
candidate descriptions through raw-frame annotation, an alternative
FFmpeg decoding and frame-sampling path, and relabeling conditioned
on intermediate seed or prior descriptions.
The FFmpeg path changes the decoding and frame-sampling implementation,
while seed- and prior-conditioned passes provide textual drafts from
earlier stages or other annotation passes. A final language-model selector compares the candidate descriptions
and selects one according to the completed-action guidelines, favoring
concrete action--object--destination/state phrasing.
The selector receives candidate texts only, not video frames; visual
evidence is incorporated through the preceding annotation passes.
We pair the selected description with its refined temporal interval
to obtain the final EgoDex subtask annotation.
The annotation configuration is summarized in
\Cref{tab:egoant-config}.

\begin{table}[t]
    \centering
    \small
    \setlength{\tabcolsep}{4pt}
    \caption{\textbf{EgoANT configuration for human-data annotation.}
    Each row summarizes the inputs and model used at one stage.
    Candidate selection uses the listed model in text-only mode.}
    \label{tab:egoant-config}
    \begin{tabular}{
        @{}
        p{0.22\linewidth}
        p{0.52\linewidth}
        p{0.18\linewidth}
        @{}
    }
        \toprule
        \textbf{Stage}
        & \textbf{Input / Configuration}
        & \textbf{Model} \\
        \midrule
        Global segmentation
        & Full-episode timestamped contact sheets;
          $0.5$\,s sampling, up to 20 frames per sheet,
          and completed-event rules
        & Qwen3.6-27B \\

        Local refinement
        & Local timestamped contact sheets, coarse hypothesis,
          and episode-level context; no additional window padding
        & Qwen3.6-27B \\

        Segment annotation
        & Fixed temporal segments; raw-frame, FFmpeg-based,
          and seed- or prior-conditioned annotation passes
        & Qwen3.5-397B \\

        Candidate selection
        & Candidate description texts only; no video frames
        & Qwen3.5-397B \\
        \bottomrule
    \end{tabular}
\end{table}

\noindent\textbf{Temporal alignment and motion standardization.}
Across all three datasets, we temporally align the subtask intervals
and descriptions with the corresponding egocentric frames,
head-camera poses, and bilateral wrist trajectories.
Each wrist pose is represented by a 3D position and a quaternion,
$[x,y,z,q_x,q_y,q_z,q_w]$.
We transform both wrist trajectories into the coordinate frame of
the head-mounted camera at the conditioning time, keeping this
reference frame fixed throughout the sampled sequence.
Concatenating the left- and right-wrist poses yields a
14-dimensional human motion representation.
We remove samples with missing annotations, invalid temporal
alignment, insufficient duration for the sampled sequence, or
implausible wrist geometry.
The resulting data share a consistent representation of visual
observations, subtask descriptions, and human motion for pretraining.

Additional implementation details, evaluations of alternative
segmentation and labeling configurations, and the complete prompt
templates used at each stage are available in the accompanying
\href{https://www.shuai-zhou.com/egoant-report/?lang=en}{EgoANT report}.

\section{Method}
\label{sec:method}

\subsection{Model Architecture}
%@家毅

\label{sec:architecture}

\begin{figure}[t]
    \centering
    \includegraphics[width=\textwidth]{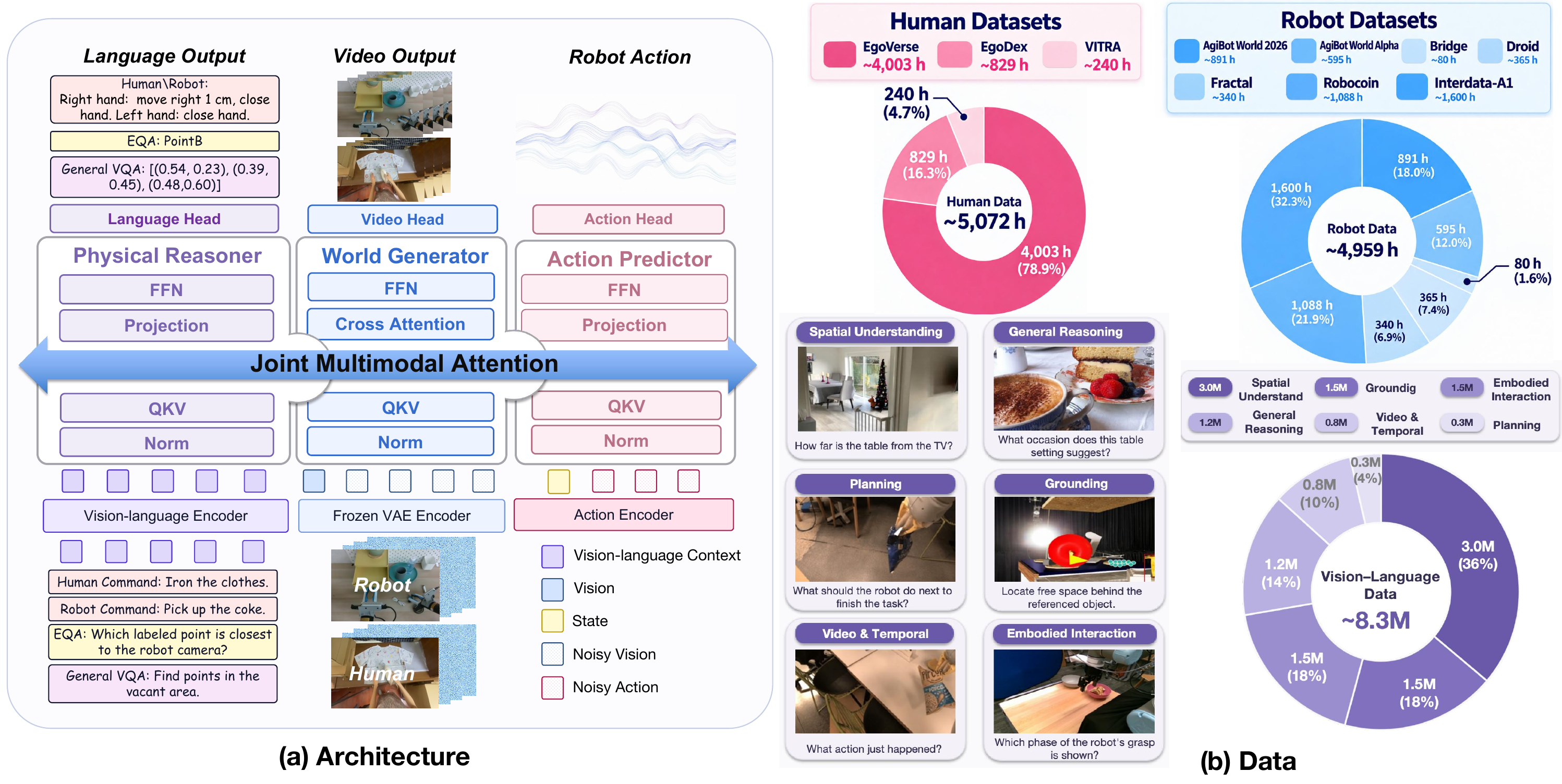}
    \vspace{-0.4cm}
    \caption{\textbf{Overview of our \method.}
    %\method由physical reasoner, world generator, and action predictor通过joint multimodal attention连结构成。robot和human egocentric data共同用于训练三个专家，其中对于physical reasoner的监督通过physical language的形式实现。此外，general vqa和eqa data也用于训练physical reasoner。
    \textbf{(a)} \method~comprises a physical reasoner, a world generator, and an action predictor interconnected through joint multimodal attention. Robot data and human egocentric data are jointly used to train all three experts, with supervision for the physical reasoner provided in the form of physical language. General VQA and Embodied Question Answering (EQA) data are additionally used to train the physical reasoner.
    \textbf{(b)} Overview of the pretraining data:
    Human and robot data are measured in hours; vision--language data are
    measured in millions of samples.
    Percentages are computed within each data category.
    }
    \label{fig:uniwam-architecture}
\vspace{-0.5cm}
\end{figure}

As shown in \Cref{fig:uniwam-architecture}, \method~unifies semantic understanding, language modeling, visual generation, and action prediction in a Mixture-of-Transformers (MoT) architecture~\citep{liang2025mot}, i.e., physical reasoner, world generator, and action predictor.
They exchange information through joint attention.
% combining semantic context, future visual dynamics, and action generation.
Given the current observation $o_t$, proprioceptive state $s_t$, and task instruction $I$, the model
learns the joint distribution of language outputs, future observations, and actions:
\begin{equation}
 p_\theta\!\left(y,o_{t+1:t+h},a_{t+1:t+h}\mid o_t,s_t,I\right),
 \label{eq:joint-prediction}
\end{equation}
where $y$ is the output language sequence of physical language tokens
or answer token.
Although \method~shares similarities with world action models such as Motus~\citep{bi2025motus} in its decoder-layer structure, it differs in its training recipe, data composition, and overall capabilities.

\noindent\textbf{Modality-specific experts.}
The physical reasoner projects features from
Qwen3-VL-2B-Instruct~\citep{bai2025qwen3vl}, conditioned on $o_t$ and $I$,
into semantic tokens and retains its language head for autoregressive
language generation, trained with the language supervision objective
in Eq.~\ref{eq:language-loss}.
The world generator uses the Wan2.2-TI2V-5B backbone~\citep{wan2025wan} to predict future
visual flow in a frozen autoencoder's latent space, with the current
observation latent kept clean.
The action predictor embeds the action chunk
with positional information and a current-state token encoding $s_t$, and predicts its
continuous flow field.
% The world generator and action predictor use separate timestep
% embeddings to accommodate different noise levels.

\noindent\textbf{Cross-modal interaction.}
Let $X_m^\ell$ denote the tokens of modality $m\in\{u,v,a\}$ at layer
$\ell$, corresponding to understanding, video, and action.
After expert-specific normalization and, for video and action, timestep
modulation, we project the tokens into a common attention space:
\begin{equation}
 Q_m=\widetilde{X}_m^\ell W_m^Q,\qquad
 K_m=\widetilde{X}_m^\ell W_m^K,\qquad
 V_m=\widetilde{X}_m^\ell W_m^V.
\end{equation}
For each attention head, the projected sequences are concatenated along
the token dimension and jointly attended:
\begin{equation}
 O=\operatorname{softmax}\!\left(\frac{QK^\top}{\sqrt{d}}+M\right)V,
 \qquad Q=[Q_v;Q_a;Q_u],
 \label{eq:joint-attention}
\end{equation}
where $K$ and $V$ follow the same concatenation order, $d$ is the head
dimension, and $M$ is the attention mask.
% During physical language supervision, the mask excludes supervised answer suffixes from joint-attention keys and values to prevent target leakage.
The outputs $O$ are split by modality and returned through expert-specific
projections, residual connections, and feed-forward networks.
This repeated interaction lets action tokens incorporate both semantic
context and evolving visual predictions while preserving modality-specific
processing throughout the network.

% 中文对照（整体架构）
% UniWAM 在单一 Mixture-of-Transformers（MoT）架构中统一语义理解、语言建模、视觉预测与动作生成。
% 采用 MoT 架构，通过联合注意力连接理解、视频和动作三路专家，融合语义上下文、未来视觉动态和动作生成。
% 各专家保留独立的 token 表征和前馈网络，并在每一层 MoT 中交换信息。
% 给定当前观测 o_t、本体状态 s_t 和任务指令 I，模型学习语言输出、未来观测与动作的联合分布 p_theta(y,o_{t+1:t+h},a_{t+1:t+h}|o_t,s_t,I)，其中 y=(y_1,...,y_T) 为包含 T 个 token 的输出语言序列，h 为预测时域。
%
% 中文对照（模态专家）
% 理解专家将 Qwen3-VL-2B-Instruct 基于 o_t 和 I 提取的特征投影为语义 token，并保留语言头用于自回归语言生成，通过语言监督目标训练。
% 视频专家采用 Wan 骨干，在冻结自编码器的潜空间中预测未来视觉流，并保持当前观测潜变量干净。
% 动作专家编码动作片段、位置信息及当前状态 token，预测连续动作流场；视频与动作专家采用独立时间步嵌入以处理不同噪声水平。
%
% 中文对照（跨模态交互）
% 每一层中，各专家的 token 经独立归一化后投影到共同的注意力空间；视频与动作路径还接受时间步调制。
% 三路查询、键和值沿 token 维拼接并计算联合注意力，再按模态拆分输出，经过独立的投影、残差连接和前馈网络更新。
% 在语言动作监督时，注意力掩码将目标答案后缀排除在联合注意力的键和值之外，避免目标信息泄漏。
% 这种逐层交互使动作 token 同时结合语义上下文与持续更新的视觉预测，并保留各模态独立的处理路径。

\subsection{Pretraining}
\subsubsection{Physical Language Supervision}
%@家毅
To preserve pretrained VLM knowledge during action supervision, we encode
end-effector motions as natural-language descriptions, following
LAP~\citep{zha2026lap}. A fixed coordinate convention and structured
templates provide deterministic descriptions of continuous actions at
arbitrary precision. We apply this supervision to both human and robot data:
task instructions specify the overall goal, while physical-language targets
describe the local behavior associated with the current observation. Given
observation $o_t$ and instruction $I$, VLM predicts target
tokens autoregressively:
\begin{equation}
  \mathcal{L}_{\mathrm{lang}}
  =-\sum_{j=1}^{T}\log p_\theta(y_j\mid o_t,I,y_{<j}),
  \label{eq:language-loss}
\end{equation}
where loss is computed only on answer tokens. Describing end-effector pose
changes provides shared motion semantics across embodiments and complements
continuous action supervision.
% By capturing coarse motion trends, this auxiliary objective
% complements the action predictor's continuous action prediction.
% As described in Section~\ref{sec:architecture}, supervised answer tokens
% are excluded from joint-attention keys and values to prevent target
% leakage into the World Generator and Action Predictor.

% 中文对照（语言动作监督）
% 参考 LAP，我们使用局部动作的自然语言描述监督理解专家。
% 在预训练中，第一人称人类数据与机器人数据均采用语言动作监督，为两类数据提供共同的监督接口。
% 任务指令描述整体目标，语言动作目标则描述当前观测对应的局部行为。
% 给定当前观测 o_t 和任务指令 I，VLM 的语言输出头以自回归方式预测目标序列 y，仅在答案 token 上计算损失。
% 末端位姿变化的语言化描述为包括人类在内的不同本体提供了共享的运动表达，促使 VLM 学习抽象于本体差异的统一内在动力学表征。
% 这一辅助监督刻画粗略的运动趋势，与动作专家的连续动作预测互为补充。
% 目标答案 token 不参与联合注意力的键和值，避免向视频和动作专家泄漏目标信息。

\subsubsection{Training Objectives}
%@家毅
Let $\hat{u}^{v}_{\theta}$ and $\hat{u}^{a}_{\theta}$ denote
the visual and action velocity fields predicted by the model
with parameters $\theta$ at the sampled noise levels
$\tau_v$ and $\tau_a$, respectively.
We supervise these predictions with flow matching losses,
using the differences between source and clean target samples
as the target velocity fields:
\begin{equation}
\resizebox{0.92\linewidth}{!}{$\displaystyle
 \mathcal{L}_{v}
 =\mathbb{E}_{\mathcal{D},\tau_a,\tau_v,A_t^1,Z_t^1}
 \!\left[\left\|\hat{u}^{v}_{\theta}-(Z_t^1-Z_t^0)\right\|_2^2\right],
 \qquad
 \mathcal{L}_{\mathrm{act}}
 =\mathbb{E}_{\mathcal{D},\tau_a,\tau_v,A_t^1,Z_t^1}
 \!\left[\left\|\hat{u}^{a}_{\theta}-(A_t^1-A_t^0)\right\|_2^2\right]
$}
\label{eq:visual-loss}
\end{equation}

where $\mathcal{D}$ is the training data distribution,
$A_t^0=a_{t+1:t+h}$ is the target action chunk, and $Z_t^0$ contains
future visual latents from the frozen video autoencoder.
The visual variables and $\hat{u}^{v}_{\theta}$ cover only future frames;
the conditioning observation $o_t$ does not contribute to the visual loss.
We independently sample action and video noise levels $\tau_a$ and $\tau_v$
from their respective training schedules and form the interpolants
$A_t^{\tau_a}=(1-\tau_a)A_t^0+\tau_a A_t^1$ and
$Z_t^{\tau_v}=(1-\tau_v)Z_t^0+\tau_v Z_t^1$.
Both velocity fields are predicted in a joint forward pass.
During pretraining, both the action source $A_t^1$ and the visual source
$Z_t^1$ are Gaussian noise.
Generation proceeds from the source ($t=1$) to the clean target ($t=0$).
The total objective combines action, visual, and language losses:
\begin{equation}
 \mathcal{L}=w_a\mathcal{L}_{\mathrm{act}}
             +w_v\mathcal{L}_{v}
             +\mathcal{L}_{\mathrm{lang}},
 \label{eq:training-objective}
\end{equation}
where $w_a$ and $w_v$ balance the action and visual objectives.
The language loss uses the autoregressive objective in
Eq.~\ref{eq:language-loss}, with the answer sequence given by either a
physical language description or a visual question answering (VQA) response.

% 中文对照（训练目标）
% hat{u}^v_theta 和 hat{u}^a_theta 分别表示参数为 theta 的模型在采样噪声水平 tau_v 和 tau_a 下预测的视觉与动作速度场；以源样本减去干净目标样本作为目标速度场。
% 动作与视觉流匹配损失采用预测流场与源减目标位移之间的平方范数，期望涵盖数据分布、两路噪声水平及源采样。
% A_t^0=a_{t+1:t+h} 为目标动作，Z_t^0 为冻结视频自编码器产生的未来视觉潜变量；视觉变量和预测流场仅覆盖未来帧，当前条件观测 o_t 不参与视觉损失。
% 两路从各自训练调度器独立采样 tau_a、tau_v，分别构造目标与源之间的线性插值，并通过一次联合前向预测两路流场。
% 预训练中的动作源和视觉源均为高斯噪声。噪声水平 1 对应源，0 对应干净目标，生成沿 1 到 0 进行。
% 总损失由动作、视觉与语言三项组成，动作项和视觉项分别使用权重 w_a 与 w_v，语言项系数为 1。
% 语言项采用前文的自回归目标，答案可以是语言动作描述或 VQA 回答，仅答案 token 参与损失。

\subsection{Posttraining}
%@家毅
In addition to maintaining supervision over all three experts
% to strengthen the model's capabilities across multiple dimensions in downstream tasks
, we further introduce 1) future-frame noise augmentation to promote information transfer between visual and action prediction, and 2) history-conditioned flow matching to improve action prediction efficiency and performance.
% We further coordinate visual and action prediction through future-frame noise augmentation and history-conditioned flow matching, respectively.
\subsubsection{Future-frame Noise Augmentation}
\label{sec:future-noise}
Joint attention can make action prediction overly dependent on nearly clean
future visual latents that are sometime unreliable at inference.
Thus, we conduct visual noise augmentation that retain joint attention and augment only future-frame latents during posttraining. 
With probability $0.5$ per sample, we further perturb the
interpolated visual input $Z_t^{\tau_v}$ as
\begin{equation}
 \widetilde{Z}_t^{\tau_v}
 =s_{\mathrm{aug}}Z_t^{\tau_v}+(1-s_{\mathrm{aug}})\epsilon_v,
 \quad s_{\mathrm{aug}}\sim\mathcal{U}(0.5,1),\quad
 \epsilon_v\sim\mathcal{N}(0,\mathbf{I}).
 \label{eq:future-augmentation}
\end{equation}
Otherwise, the input is unchanged. 
The current observation latent remains
clean, and the original flow targets and timestep embeddings are retained.
This input perturbation encourages action prediction to tolerate imperfect
future visual representations while preserving the observed task context.

% 中文对照（未来帧噪声增强）
% 联合注意力可能使动作预测过度依赖接近干净的未来视觉潜变量，而这些信息在推理时可靠性较低。
% 我们参考 LingBot-VA，使用视觉噪声增强。
% 我们保留联合注意力，在后训练中以每个样本 0.5 的概率，对流匹配插值后的未来视觉潜变量额外混入高斯噪声。
% 保留系数 s_aug 从 [0.5,1] 均匀采样；未选中的样本保持不变。
% 该输入扰动旨在使动作预测适应不完美的未来视觉表征，同时保留已观测的任务上下文。

\subsubsection{History-conditioned Flow Matching}
\label{sec:history-flow}
\begin{wrapfigure}{r}{0.52\textwidth}
    \centering
    \includegraphics[width=\linewidth]{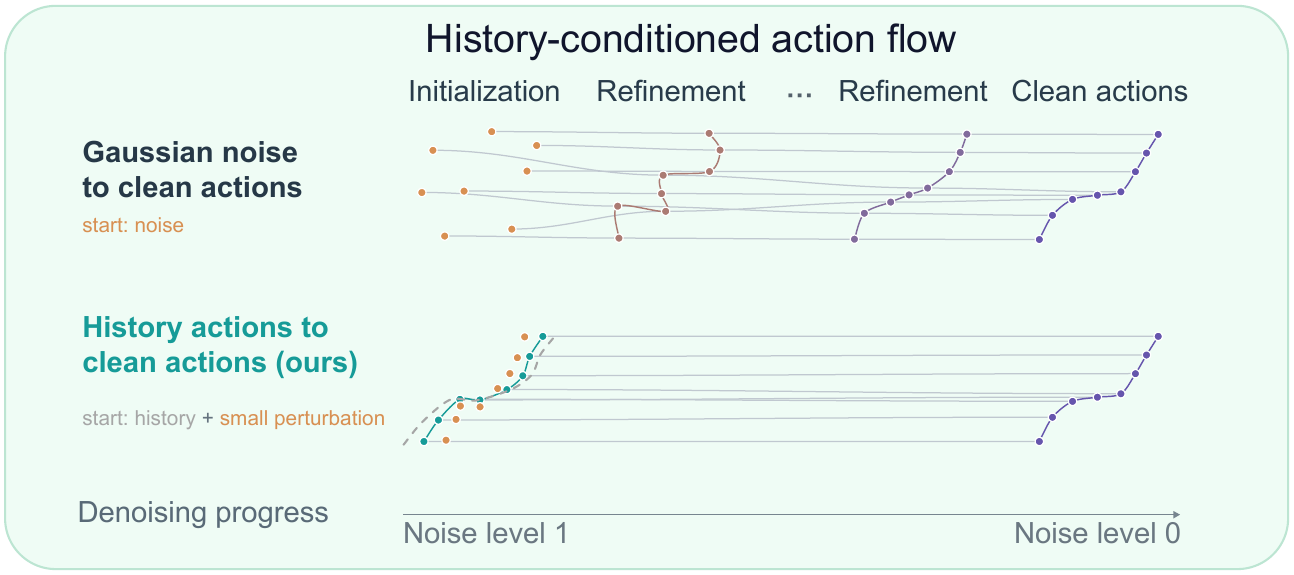}
    \caption{\textbf{History-conditioned action denoising.} Action generation starts from perturbed action history rather than Gaussian noise.}
    \label{fig:a2a}
\end{wrapfigure}
Recent actions provide a structured prior for predicting subsequent actions~\citep{action2actionflow}.
As illustrated in \Cref{fig:a2a}, for posttraining, we replace the Gaussian noise source of the action flow
with a perturbed chunk of previously executed actions, grounding generation
in action history to encourage temporal consistency and support
refinement with fewer inference steps. The visual source remains Gaussian noise.
Let $A_t^0=a_{t+1:t+h}$ denote the target action chunk and
$A_{t-1}=a_{t-h+1:t}$ the corresponding history of $h$ actions, sampled at
the action rate. Thus, $A_{t-1}$ and $A_t^0$ share the same dimensions. We
construct the source as
\begin{equation}
 A_t^1=A_{t-1}+\epsilon_a,\qquad
 \epsilon_a\sim\mathcal{N}(0,\eta^2\mathbf{I}),
 \label{eq:history-source}
\end{equation}
where $\eta=0.02$ introduces small stochastic perturbations and
$\mathbf{I}$ denotes the identity matrix.

We write $A_t^\tau$ for the action chunk at normalized noise level
$\tau\in[0,1]$, with $t$ indexing control time. The interpolation
between the clean target and history source is:
\begin{equation}
 A_t^\tau=(1-\tau)A_t^0+\tau A_t^1,\qquad
 u_t^a=A_t^1-A_t^0.
 \label{eq:flow-path}
\end{equation}
The action predictor learns this displacement with a mean-squared flow
matching loss conditioned on the current context, while exchanging features
with the world generator through joint attention.

%% file: table/dataset_durations.tex
\begin{table}[H]
    \centering
    \caption{Overview of the training datasets.}
    \label{tab:dataset_durations}
    \begin{tabular}{lllr}
        \toprule
        Data Type & Embodiment Type & Data Sources & Duration (hours) \\
        \midrule
        Robot & Dual-arm/Mobile & AgiBot World 2026 & $\sim$891 \\
         & Dual-arm/Mobile & AgiBot World Alpha & $\sim$595 \\
         & Single-arm & Bridge & $\sim$80 \\
         & Single-arm & Droid & $\sim$365 \\
         & Single-arm & Fractal & $\sim$340 \\
         & Dual-arm & Robocoin & $\sim$1088 \\
         & Dual-arm/Mobile & Interdata-A1 & $\sim$1600 \\
        \midrule
        \multicolumn{3}{l}{Robot subtotal} & $\sim$4958 \\
        \midrule
        Human & Human hands & EgoVerse & $\sim$4003 \\
         & Human hands & EgoDex & $\sim$829\\
         & Human hands & VITRA & $\sim$240\\
        \midrule
        \multicolumn{3}{l}{Human subtotal} & $\sim$5072 \\
        \midrule
        \multicolumn{3}{l}{Total} & $\sim$10013 \\
        \bottomrule
    \end{tabular}
    % Robot subtotal is rounded from the original durations (4363.06 hours).
\end{table}
%@周帅

%% file: table/vqa_data.tex
\begin{table}[t]
    \centering
    \caption{Composition of the vision--language pretraining data.}
    \label{tab:vqa_composition}

    \small
    \setlength{\tabcolsep}{4.2pt}
    \renewcommand{\arraystretch}{0.94}

    \begin{tabular}{llr}
        \toprule
        \textbf{QA Type}
        & \textbf{Data Source}
        & \textbf{QA Pairs} \\
        \midrule

        \textbf{Spatial Understanding}
        & SAT
        & 172K \\
        & RefSpatial
        & 1.430M \\
        & VST-P
        & 563K \\
        & SenseNova-SI
        & 800K \\
        & GRiD-3D
        & 358K \\

        \cmidrule(lr){1-3}

        \textbf{Grounding}
        & RoboPoint
        & 930K \\
        & RefSpatial
        & 570K \\

        \cmidrule(lr){1-3}

        \textbf{General Reasoning}
        & CLEVR
        & 700K \\
        & RoboPoint
        & 500K \\

        \cmidrule(lr){1-3}

        \textbf{Video \& Temporal}
        & VSI-590K
        & 591K \\
        & SIMS-VSI
        & 203K \\

        \cmidrule(lr){1-3}

        \textbf{Embodied Interaction}
        & RoboVQA
        & 636K \\
        & Robo2VLM
        & 540K \\

        \cmidrule(lr){1-3}

        \textbf{Planning}
        & RoboVQA
        & 162K \\
        & Robo2VLM
        & 138K \\

        \midrule
        \multicolumn{2}{l}{\textbf{Total Vision--Language Data}}
        & \textbf{8.293M} \\
        \bottomrule
    \end{tabular}
\end{table}

%% file: sec/04_experiments.tex
\section{Experiments}
\label{sec:experiments}
We design our experiments to answer the following questions:
% @wenxuan
% 实验部分主要参考G05, LAP和flex-pi
%要把ood，few-shot，long-horizon这几个事情讲好
%有性能但并非sota的全都放进tech report
\textbf{(Q1)} Can our \method~successfully perform manipulation tasks under in-domain conditions, complex perturbations, and OOD conditions, respectively? (\Cref{sec:4.1})
% \textbf{(Q2)} Does our \method~hold that performance under OOD conditions? (\Cref{sec:4.1})
\textbf{(Q2)} Does the unified design of \method~improve instruction following and long-horizon manipulation in the real world? (\Cref{sec:4.2})
\textbf{(Q3)} Does our unified model scale effectively with data? (\Cref{sec:4.3})
\textbf{(Q4)} How does our physical language shape the learned representations and enable strong generalization? (\Cref{sec:4.3})
\textbf{(Q5)} How important are our \method’s training designs in policy performance? (\Cref{sec:4.5})

\subsection{Simulated Experiments}
\label{sec:4.1}
% We mainly conduct the simulated experiments in LIBERO~\citep{liu2023libero} and RoboTwin 2.0~\citep{chen2025robotwin2}, as well as their OOD variants, LIBERO-Plus~\citep{fei2025liberoplus} and RoboTwin 2.0 Clean2rand~\citep{community2026xpolicylab}.
We mainly conduct the simulated experiments in LIBERO~\citep{liu2023libero} and RoboTwin 2.0 Clean2Clean~\citep{chen2025robotwin2}, as well as their OOD variants, LIBERO-Plus~\citep{fei2025liberoplus} and RoboTwin 2.0 Clean2rand~\citep{community2026xpolicylab}.
\input{table/libero_results}
\input{table/robotwin_summary}

\noindent\textbf{In-distribution Evaluation.}
%我们分别在libero和libero-plus上进行indomain的模型评估。libero （cite） 包含四个suite：LIBERO-Spatial,LIBERO-Object,LIBERO-Goal,andLIBERO-Long，每个suite分别有10个任务。而libero-plus则在libero的基础上引入七个维度的扰动：background textures, camera viewpoints, language instructions, lighting conditions, object layouts, robot initial states, and sensor noise。
%我们分别在这两个benchmark上进行对应数据集的训练并且汇报结果。数据和训练的细节见附录。
% \begin{table}[t]
% \centering
% \begin{minipage}[t]{0.49\textwidth}
% \input{table/libero_results}
% \end{minipage}\hfill
% \begin{minipage}[t]{0.49\textwidth}
% \input{table/robotwin_summary}
% \end{minipage}
% \end{table}
% We evaluate \method~on both LIBERO~\citep{liu2023libero} and LIBERO-Plus~\citep{fei2025liberoplus} under their respective training distributions. 
LIBERO~\citep{liu2023libero} and RoboTwin~2.0 Clean2Clean (C2C) are in-distribution benchmarks, where C2C denotes training on clean demonstrations and evaluating the model on clean settings.
%如图所示~\ref{tab:libero}，\method~在libero的平均成功率达到了99.2%，比之前的最好基线，Xiaomi-Robotics-0超出了0.5个点。这结果表明\method~在original task distribution具有不错的基础任务解决能力。
As shown in \Cref{tab:libero}, \method~achieves an average success rate of 99.2\% on the standard LIBERO benchmark, outperforming the previous best-performing baseline, Xiaomi-Robotics-0, by 0.5\%.
This result demonstrates that \method~maintains strong in-distribution manipulation performance.
\Cref{tab:robotwin-summary} shows \method~also achieves a high success
rate of 75.14\% in RoboTwin 2.0 C2C. 
The C2C setting demonstrates the model's bi-manual performance with limited training data.

\noindent\textbf{OOD Evaluation for Robustness and Generalization.}
LIBERO-Plus~\citep{fei2025liberoplus} extends LIBERO with systematic variations along seven dimensions, including background textures, camera viewpoints, language instructions, lighting conditions, object layouts, robot initial states, and sensor noise.
As shown in Table~\ref{tab:libero_plus}, \method~achieves an overall success rate of 92.6\% on LIBERO-Plus, the highest reported value among the listed methods.
At the dimension level, \method~has the highest reported success rates among the listed methods under robot (89.5\%), language (92.2\%), light (97.9\%), and background (97.6\%) perturbations.
These results are consistent with the intended roles of our training objectives: physical language supervision encourages a shared mapping between semantics and actions, reducing reliance on spurious visual cues; pretraining with VQA data helps preserve language understanding; and large-scale world generation training encourages the model to capture underlying physical dynamics that remain consistent across variations in lighting and background.

In RoboTwin~2.0 Clean2Rand (C2R)~\citep{chen2025robotwin2}, models are trained using clean data but tested with domain randomization; \method~achieves a success rate of 68.32\%, the highest reported value among the methods listed in \Cref{tab:robotwin-summary}.
Its success rate decreases by only 6.82 percentage points from C2C to C2R, compared with 24.70 and 50.46 percentage points for $\pi_{0.5}$ and Spatial Forcing, respectively. 
These results demonstrate that \method~capabilities generalize to different heights, distractors, and backgrounds, which benefit from the physical reasoning and dynamics understanding that are unified in our hybrid model.

\input{table/libero_plus_results}

\subsection{Real-world Experiments}
\label{sec:4.2}
\begin{figure}[!t]
    \centering
    \includegraphics[width=0.8\linewidth]{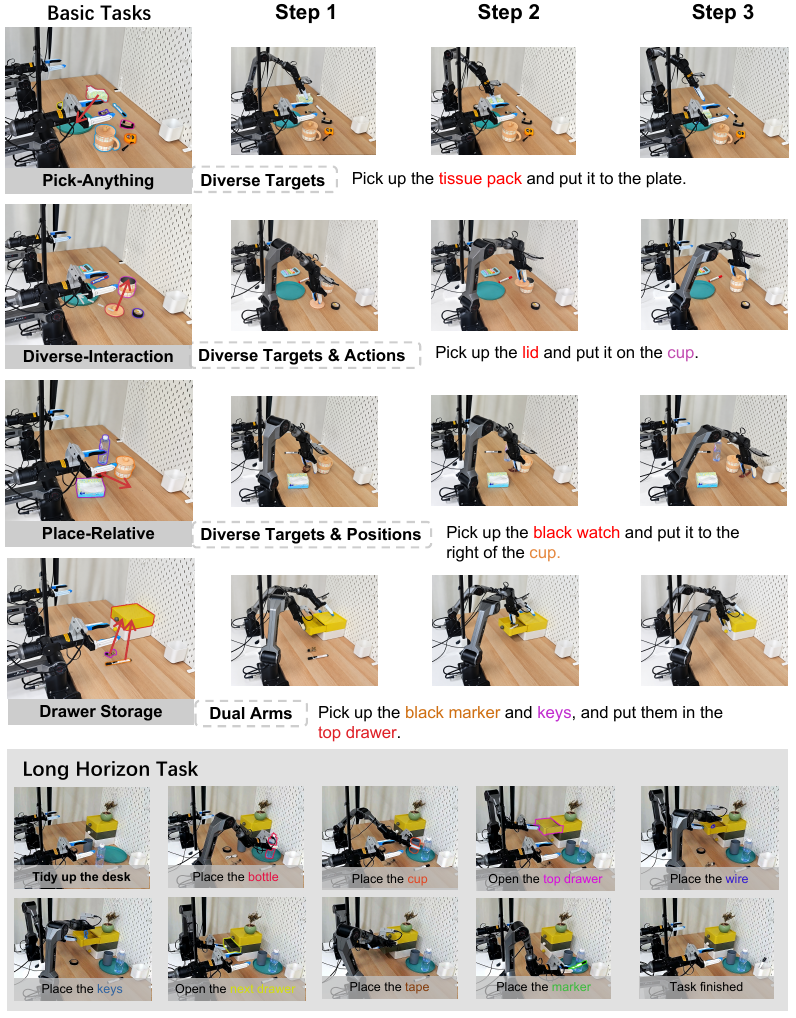}
    \caption{\textbf{Overview of our real-world evaluation tasks.}
The top panel shows representative execution sequences for four instruction-following tasks:
Pick-Anything, Diverse-Interaction, Place-Relative, and Drawer Storage.
The bottom panel presents a long-horizon task,
where the robot sequentially completes multiple manipulation stages.}
    \label{fig:manipulation_real_task}
\end{figure}
%在真实世界实验，我们注重探索模型的语言跟随能力以及长序任务解决能力，我们的实验物体多达18个，涉及不同的颜色、构型和对应的交互方式。我们使用agilex的piper开展实验，该机械臂具有6-DoF 关节，以及一个夹爪。我们的真机系统采用两个pipers，它们各自配有一个腕部的摄像头，中间高处还有个俯视的摄像头。我们我们将模型与Motus和pi05进行比较，分别代表wam基线和vla基线。所有模型都在8张h100上训练80k，每种任务我们采集200条数据，所有任务混在一起训出一个通用模型。
In our real-world experiments, we focus on evaluating the model's instruction-following ability and long-horizon task-solving capability. Our setup involves up to 18 objects with diverse colors, geometries, and associated interaction patterns. Experiments are conducted on the AgileX Piper platform, where each arm has 6-DoF joints and a parallel gripper. Our dual-arm system consists of two Piper, each equipped with a wrist-mounted camera, together with an additional third-person camera providing a global view of the workspace.

We compare \method~against Motus and $\pi_{0.5}$, representing WAM-based and VLA-based baselines, respectively. For all methods, we collect 200 demonstrations per task and jointly train a single generalist policy over all tasks. All models are trained for 80K steps on 8 NVIDIA H100 GPUs under the same training protocol.

\noindent\textbf{Instruction-following Tasks.}
%我们设计了四个任务来来考察模型的语言跟随能力：1.Pick-Anything 多物体抓取任务，考察模型的名词分辨能力； 2. Diverse-Interaction 多物体不同交互实验，考察模型的名词和动词分辨能力；3. Place-Relative 抓取物体放置到另一不同物体的不同方位实验，考察模型的名词和介词分辨能力；4. pull drawer， 把物体收纳到抽屉里，考察双臂协作和空间理解能力；
To evaluate instruction-following capability, we design four complementary tasks that require the policy to ground different linguistic components into appropriate manipulation behaviors.
As shown in Fig.~\ref{fig:manipulation_real_task}, we consider four tasks.
\textbf{(1) Pick-Anything} requires the robot to identify and grasp a target object from a multi-object scene, primarily evaluating noun grounding and object discrimination.
\textbf{(2) Diverse-Interaction} requires the robot to perform different interactions on multiple objects according to the instruction, jointly testing noun and verb grounding.
\textbf{(3) Place-Relative} requires the robot to pick a specified object and place it at a designated spatial relation to another object, evaluating noun grounding and spatial-relation understanding, particularly the interpretation of prepositions.
\textbf{(4) Drawer-Storage} requires the dual-arm system to manipulate a drawer and place the target object inside, testing bimanual coordination, spatial reasoning.
%对于这些任务，我们汇报两个指标：语言遵循率指的是模型是否抓取或操作恰当的物体比如指令抓取粉色杯子，不能去抓橙色杯子等.任务成功率对应任务是否成功执行。具体的任务细节见附录

For these tasks, we report two complementary metrics: Instruction-Following Rate (IFR) and Task Success Rate (SR). IFR measures whether the policy correctly grounds the language instruction to the intended object and interaction. For example, given the instruction “place the pink cup on the plate,” manipulating the orange cup is considered an instruction-following failure, irrespective of whether the subsequent manipulation is successfully executed. In contrast, SR measures whether the instructed task is successfully completed in its entirety.  Detailed task configurations and evaluation criteria are provided in the Appendix.~\ref{app:real-world}.
\begin{figure}[t]
    \centering
    \includegraphics[width=\linewidth]{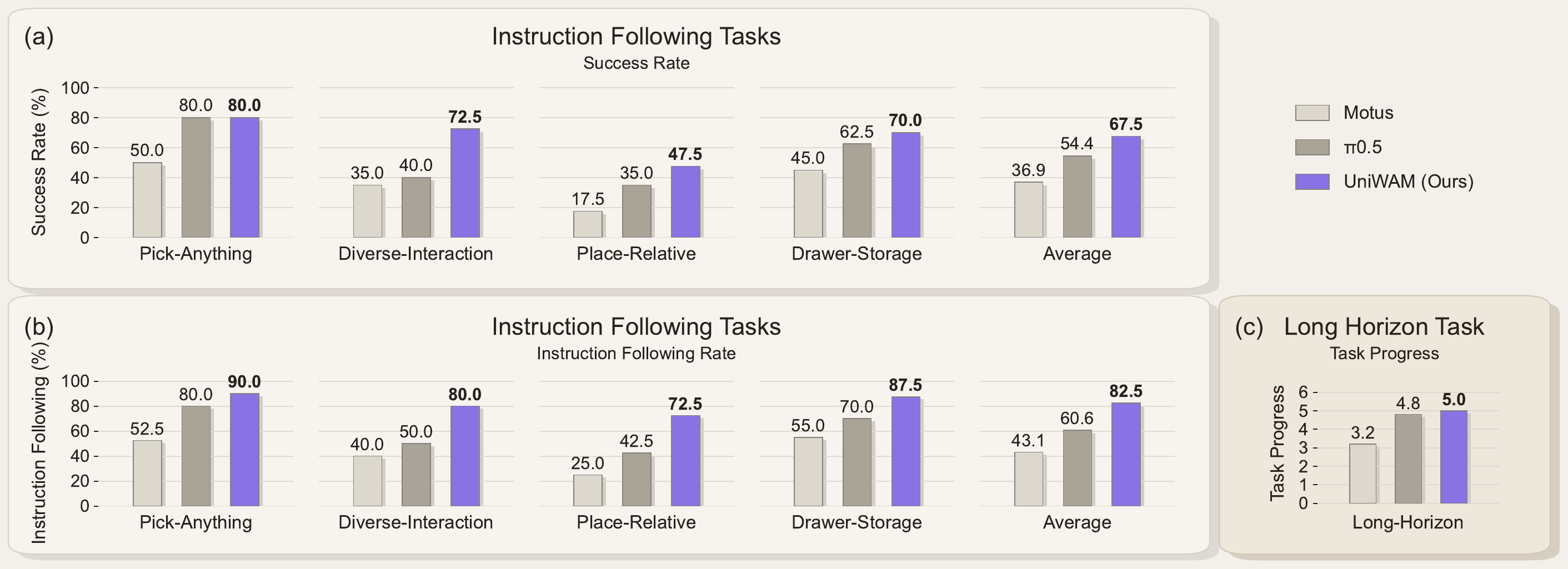}
    \caption{Real-world instruction-following evaluation across four manipulation tasks.}
    \label{fig:basic_real_task}
\end{figure}
%如图所示,我们的method在across tasks with different levels of linguistic and manipulation complexity不管是语言跟随率和任务成功率都实现了不错的表现.我们的method achieves a success rate of 67.5% and an instruction-following rate of 82.5%, improving over pi0.5 by 13.1 and 21.9 percentage points, respectively.优势在language grounding上比较明显.特别是在Diverse-Interaction,\method improves the success rate from 40.0% to 72.5% and the instruction-following rate from 50.0% to 80.0%.这些都说明了我们的language–action supervision帮助模型更好通过composite instruction更好地grounding 恰当地行为.

As shown in Fig.~\ref{fig:basic_real_task}(a,b), \method~consistently achieves strong performance across tasks with varying levels of linguistic and manipulation complexity, in terms of both task success and instruction following. Averaged over the four tasks, \method~achieves a success rate of 67.5\% and an instruction-following rate of 82.5\%, outperforming \(\pi_{0.5}\) by 13.1 and 21.9 percentage points, respectively. The improvement is particularly pronounced in language grounding. On Diverse-Interaction, where the policy must jointly identify the target object and infer the intended interaction, \method~improves the success rate from 40.0\% to 72.5\% and the instruction-following rate from 50.0\% to 80.0\%. These results suggest that our physical language supervision facilitates more reliable grounding of compositional instructions into appropriate manipulation behaviors.
%这里是否再补一些分析,instruction和success之间的gap? 强调我们的理解能力ok

\noindent\textbf{Long-horizon Tasks.}
%我们设计了一个包含多个子任务的桌面整理任务来测试模型的长序任务解决能力：1. 将杯子收纳到盘子里 2. 将日常用品收纳到第一层柜子 3. 将工具收纳到第二层柜子 4. 将笔放入卡槽。单纯汇报成功率难以衡量模型对子任务的完成程度, 我们采用Task Progress指标来衡量模型的能力.从1到6依次指的是成功收纳第一个杯子, 成功收纳第二个杯子, 拉出抽屉, 将东西放进去,将抽屉关闭,将笔收纳到笔筒里.
As illustrated in Fig.~\ref{fig:manipulation_real_task}, we design a multi-stage tabletop organization task to evaluate long-horizon manipulation, involving cup placement, drawer storage, and marker placement. Since binary success cannot capture partial completion, we report Task Progress, a score from 0 to 6 defined by six sequential milestones: placing the first cup, placing the second cup, opening the drawer, placing the target object inside, closing the drawer, and placing the marker into its holder. A higher score indicates that the policy successfully progresses through more stages of the long-horizon task.
%如图所示, 我们的模型的score达到了5.0, 领先已有基线. 我们输入的文本指令仅仅只是"Tidy up the desk."并不是复杂的全指令,只在训练中在physical reasoner的文字化动作输出前加上细分任务描述,作为同样的预测目标.

As shown in Fig.~\ref{fig:basic_real_task}(c), \method~achieves an average Task Progress of 5.0, compared with 4.8 for \(\pi_{0.5}\) and 3.2 for Motus, indicating stronger sustained execution over long-horizon manipulation sequences. Notably, the policy receives only the high-level instruction “Tidy up the desk.” at inference time, without an explicit step-by-step description of the required procedure. The model must infer the relevant intermediate subgoals from the scene and its current execution state and progressively complete them over the course of the task.
During training, we prepend a fine-grained subtask description to the textual action reasoning produced by the physical reasoner and optimize both as prediction targets. This supervision encourages the model to associate a high-level task objective with intermediate subgoals and their corresponding physical actions. 

\subsection{In-depth Analysis of Human-robot Co-training}
% \subsection{Scaling Law of Robot and Human Data}
\label{sec:4.3}

\noindent\textbf{Data scaling law emerges.}
We study how the scale of robot pretraining data affects downstream OOD manipulation performance, and effects brought by the added egocentric human data. We pretrain model using 250, 500, 1k, 2.5k, and 5k hours of robot data, with alternative same amount of human egocentric data.

\begin{figure}[t]
    \centering
    \includegraphics[width=\linewidth]{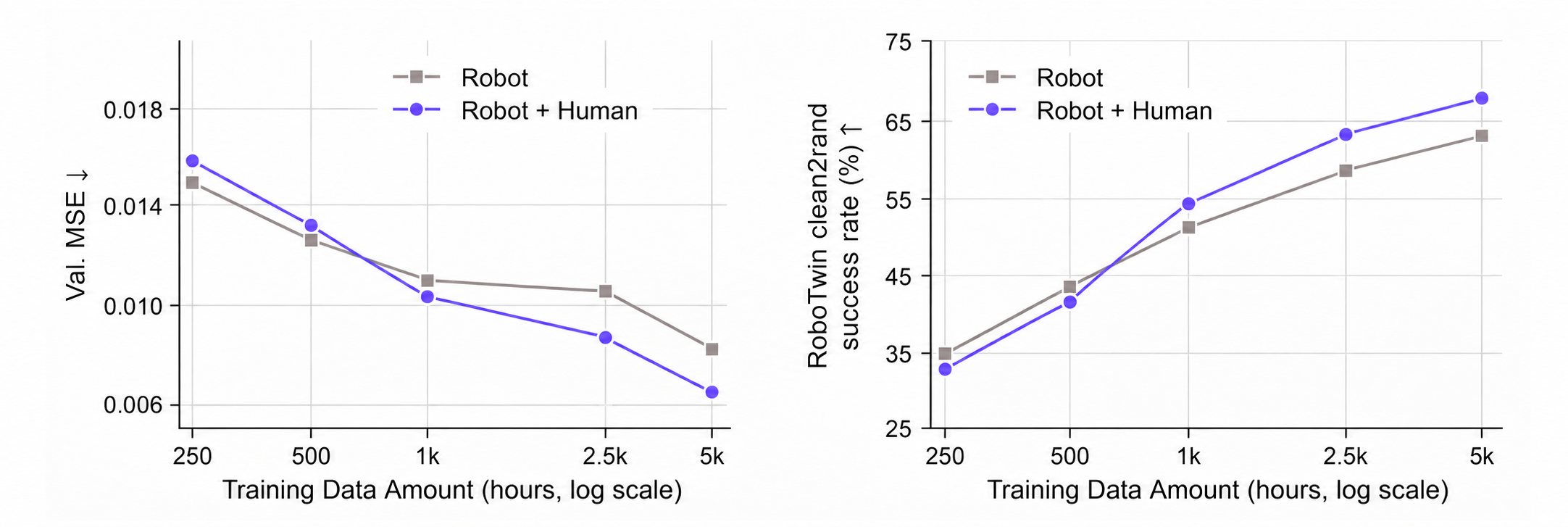}
    \caption{
    Data scaling comparison between Robot and Robot + Human. Left: validation MSE. Right: RoboTwin clean2rand success rate. The x-axis reports the number of robot-data hours on a logarithmic scale. Robot + Human augments the robot data with a fixed proportion of human data. 
    At the final data point, the training set comprises 4958 hours of robot data and 5072 hours of human data.
    All settings use mixed training with VQA data.
    }
    \label{fig:scaling-law}
\end{figure}

As shown in \Cref{fig:scaling-law}, increasing the amount of robot data leads to consistent and substantial gains in OOD performance.
Average task completion rises monotonically from 0.35 at 250 hours to 0.67 at 5k hours, with no signs of saturation in the explored regime.
At small data scales, incorporating egocentric human data degrades performance. However, once the training data reaches 1k hours, human data begins to yield gains, which become more pronounced as the dataset grows. This pattern suggests that visual and physical differences between human and robot data at small data scales hinder learning across the two distributions, whereas large-scale data may enable the transfer of shared world dynamics across embodiments, leading to improved generalization.

%值得注意的是，当我们将收敛时达到的最优验证损失与数据规模进行比较时，观察到了一个非常清晰的对数—线性缩放定律：

Furthermore, when we plot the validation loss achieved at convergence against data scale, we observe a remarkably clean log-linear scaling law:
\begin{equation}
    L(D) \approx 0.03205 - 0.00302 \ln(D), 
\end{equation}
%其中，$D$ 表示Robot + Human数据的小时数。拟合曲线的决定系数达到 $\qquad R^2 \approx 0.983.$，说明在对数空间中二者几乎呈完美的线性关系。更重要的是，这种离线缩放行为能够强有力地预测真实机器人的性能。随着数据规模变化，人类验证损失与下游任务完成度保持紧密同步，这表明验证损失不仅仅是一个离线指标，也能够反映具身控制能力。总的来说，这些结果表明，有效的人类到机器人迁移从根本上说是一个规模扩展问题。在本文探索的范围内，增加人类数据能够带来可预测的验证损失下降，以及相应的机器人性能提升；同时，没有观察到收益递减的证据。
where $D$ denotes the number of hours of robot and human pretraining data. 
The fitted curve achieves an $R^2 \approx 0.983.$, indicating an almost perfect linear relationship in log space. 
Moreover, this offline scaling trend closely tracks real-world robot performance: as the dataset grows, validation loss on pretraining data decreases alongside improvements in downstream task success rates. 
This correspondence suggests that validation loss can serve as an indicator of embodied control capability. 
Taken together, these results suggest that data scale plays a central role in effective human--robot transfer. 
% Within the range studied, incorporating more human data is associated with predictable reductions in validation loss and corresponding improvements in robot performance.

\begin{figure}[t]
    \centering
    \includegraphics[width=\linewidth]{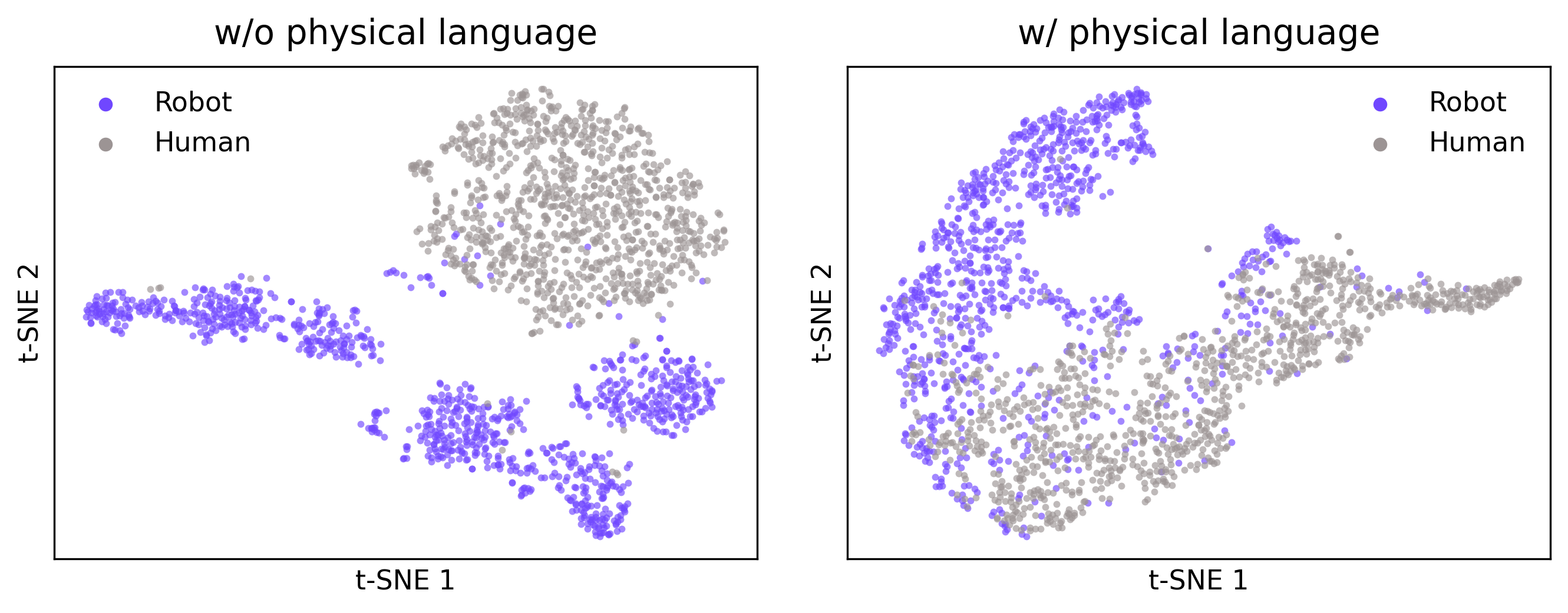}
    \caption{\textbf{Human--robot transfer emerges with physical-language pretraining.}
    t-SNE visualization of physical reasoner representations from robot (purple)
    and human (gray) data without (left) and with (right) physical language
    supervision. The representations exhibit greater overlap with supervision.}
    \label{fig:tsne-alignment}
\end{figure}

\noindent\textbf{Human-robot transfer emerges with physical-language pretraining.}
Figure~\ref{fig:tsne-alignment} shows that, with supervision, physical reasoner's representations from robot and human data exhibit greater overlap, particularly in the central and lower regions of the projection.
Without this supervision, robot and human samples largely occupy separate regions.
These results support the role of physical language supervision in reducing domain gap in the learned representations.
This pattern is attributed to shared physical language targets encouraging representations of common manipulation semantics across human and robot embodiments, thereby facilitating the general manipulation capability and improved generalization afforded by mixed-data pretraining.

% 中文对照（跨本体表征对齐）
% 图中比较了有无 LAP 式物理语言监督时机器人与第一人称人类数据的 VLM 表征。
% 未加入监督时，两类样本主要分布于不同区域，混合较少；加入监督后，投影中央和下方区域的混合更加明显，但仍保留部分来源特定区域。
% 这一现象与共享物理语言目标促进跨本体共同操作语义表征的解释一致，为减小表征中的数据来源分离提供定性支持。
% 该可视化本身不能量化原始特征空间中的对齐程度，也不能单独确定其对下游任务性能的影响。

\subsection{Ablation Studies}
\label{sec:4.5}

\begin{figure}[t]
    \centering
    \includegraphics[width=0.95\textwidth]{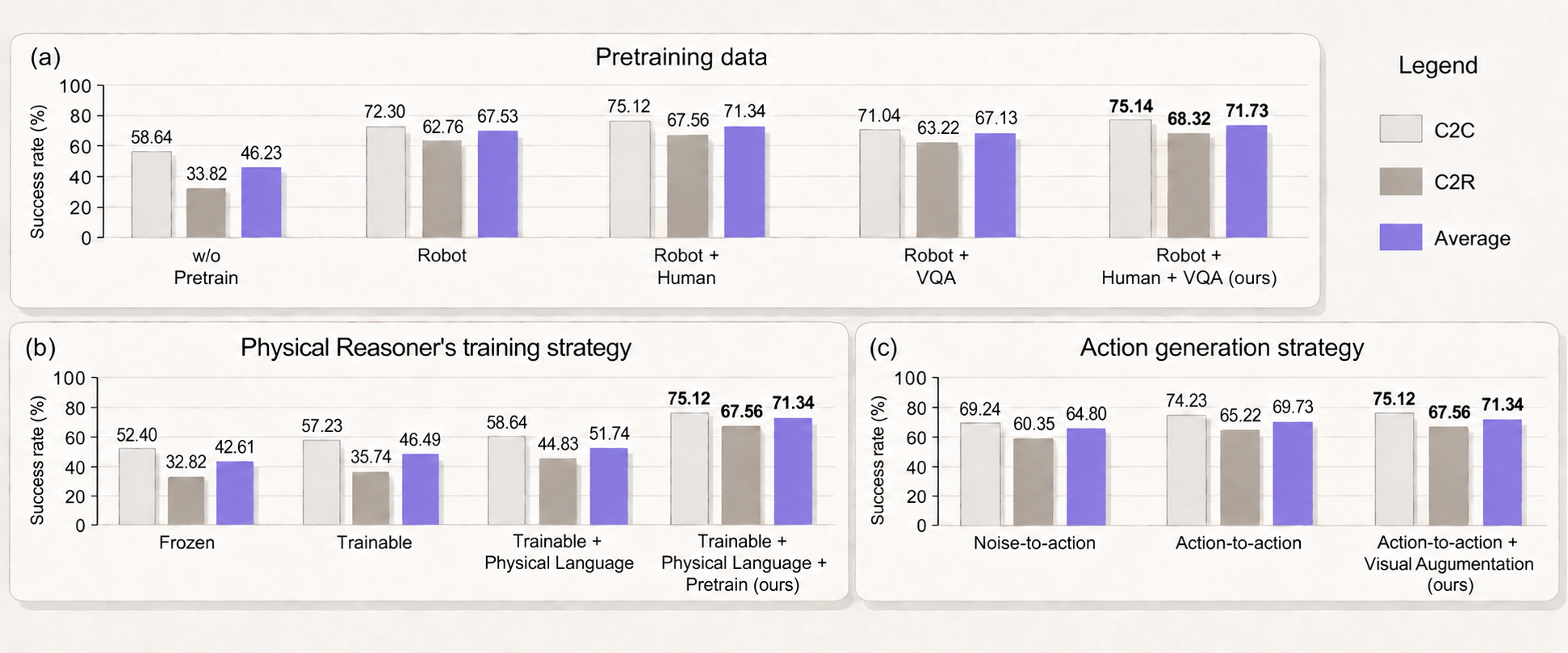}
    \caption{\textbf{Ablation studies on RoboTwin 2.0.}
    % (a) Pretraining data composition. (b) Physical reasoner's training strategy. (c) Action generation strategy.
    % C2C and C2R denote clean2clean and clean2rand; Average is their
    % arithmetic mean.
    % Fro. and Unfro. denote frozen and unfrozen VLMs, respectively.
    % PL denotes physical language supervision, and Pre. denotes pretraining.
    % N2A and A2A denote noise-to-action and action-to-action generation; Vis. Aug. denotes
    % future visual augmentation.
    }
    \label{fig:ablations}
\end{figure}
\noindent\textbf{Multi-source pretraining.}
Figure~\ref{fig:ablations}(a) compares different pretraining data compositions.
Robot-only pretraining improves the average success rate from 46.23\% to 67.53\%.
Adding human egocentric data further increases C2C success from 72.30\% to 75.12\% and C2R success from 62.76\% to 67.56\%, raising the average to 71.34\%.
By comparison, Robot + VQA achieves an average success rate of 67.13\%, with a modest improvement in C2R but a decrease in C2C relative to robot-only pretraining.
Adding VQA to the robot--human mixture yields the highest reported success rates among the evaluated variants: 75.14\% in C2C, 68.32\% in C2R, and 71.73\% on average.
These comparisons support the benefit of human egocentric data for mixed-data pretraining, with VQA providing a further improvement when combined with both robot and human data.
% These results support using complementary data sources in the pretraining mixture.
% As shown in Figure~\ref{fig:ablations}(a), the three-source combination
% achieves the highest C2C, C2R, and average success rates among the evaluated
% configurations: 75.14\%, 68.32\%, and 71.73\%, respectively.
% Robot pretraining alone raises the average from 46.23\% to 67.53\%, while
% incorporating human data further increases it to 71.34\%.
% VQA complements this robot--human combination, further improving performance
% in both evaluation settings.
% Physical language supervision is central to effective use of the combined robot and human data: removing it from Robot + Human reduces C2R success from 67.56\% to 56.40\%, a drop of 11.16 percentage points.
% Together, these results support our multi-source pretraining design, with
% shared physical language targets providing an effective supervision interface
% for cross-embodiment learning.
%这些结果表明我们的模型充分利用了不同类型数据，使得模型可以从广泛数据源中进行scale并且得到comprehensive的收益
% These results demonstrate that our model effectively leverages diverse data types, enabling it to scale with data from a broad range of sources and achieve broad performance gains.

% Source tables are retained in table/; all results are shown in Figure~\ref{fig:ablations}.

\noindent\textbf{VLM's training strategy.}
%通过物理语言监督和预训练对物理推理器进行适配，可显著提升下游控制性能。图~\ref{fig:ablations}(b) 中的累积消融实验表明，解冻推理器后，平均成功率从 42.61% 提升至 46.49%。进一步加入物理语言监督后，平均成功率提升至 51.74%，其中 C2R 成功率从 35.74% 提升至 44.83%。加入预训练后，平均成功率达到 71.34%，由此形成的完整训练方案同时提升了模型在干净环境中的表现及其向随机化环境迁移的能力。这一逐步提升的结果支持通过物理语言目标和广泛的预训练经验对预训练语义表征进行适配，从而增强其在机器人操作任务中的实用性。
% Adapting the physical reasoner with physical language supervision and pretraining substantially improves downstream control.
Training the VLM matters.
The cumulative ablation in Figure~\ref{fig:ablations}(b) shows
that unfreezing the reasoner increases average success from 42.61\% to 46.49\%.
Adding physical language supervision further raises the average to 51.74\% and adding pretraining brings the average to 71.34\%. 
% completing a training recipe that improves both clean-environment performance and transfer to randomized environments.
% This progression supports adapting pretrained semantic representations
% through physical language targets and broad pretraining experience to
% strengthen their utility for robotic manipulation.
%这证明我们的训练recipt是有效的，并且能够解锁pretraining的益处。
This demonstrates that our training recipe is effective and unlocks the benefits of pretraining for VLM.

\noindent\textbf{Action generation strategy.}
% 基于历史的动作初始化与未来视觉增强进一步提升了策略性能。如图~\ref{fig:ablations}(c) 所示，动作到动作（A2A）生成优于噪声到动作（N2A）生成，将平均成功率从 64.80% 提升至 69.73%，提高了 4.93 个百分点。未来视觉增强进一步将 C2R 成功率从 65.22% 提升至 67.56%，并使平均成功率达到 71.34%。这些结果表明，基于机器人执行历史的动作生成与训练阶段的未来视觉输入增强是策略中相互补充的两个组成部分，能够共同提升操作性能以及环境变化下的泛化能力。
History-based action initialization and future visual augmentation further improve policy performance.
As shown in Figure~\ref{fig:ablations}(c), action-to-action
generation outperforms noise-to-action, increasing average SR
from 64.80\% to 69.73\%.
Future visual augmentation further brings the average SR to 71.34\%.
Further comparisons of A2A and N2A generation in terms of success rate and denoising-step count are provided in Appendix~\ref{app:action-generation-steps}.
% These improvements support grounding action generation in the robot's
% execution history and augmenting future visual inputs during training
% as complementary components of our policy, strengthening manipulation
% performance and generalization under environmental variation.

%% file: table/libero_results.tex
% Standalone benchmark table.
\begin{table}[t]
\centering
\caption{Success rate (\%) on the standard LIBERO suites.}
\small
\label{tab:libero}
\setlength{\tabcolsep}{4.0pt}
\renewcommand{\arraystretch}{0.9}
\resizebox{0.7\linewidth}{!}{%
\begin{tabular}{lcccc|c}
\toprule
Method & Spatial & Object & Goal & Long & Average $\uparrow$ \\
\midrule
% Qwen-VLA-Base & -- & -- & -- & -- & 90.8 \\
% InternVLA-M1~\citep{ye2026st4vla} & 98.0 & 99.0 & 93.8 & 92.6 & 95.9 \\
% Wall-OSS-0.5 & -- & -- & -- & -- & 96.5 \\
$\pi_0$~\citep{black2024pi0} & 98.0 & 96.8 & 94.4 & 88.4 & 94.4 \\
PD-VLA~\citep{song2025pd} & 95.5 & 96.7 & 94.9 & 91.7 & 94.7 \\
$\pi_{0.5}$~\citep{physicalintelligence2025pi05} & 98.8 & 98.2 & 98.0 & 92.4 & 96.9 \\
GR00T-N1.7~\citep{nvidia2026gr00tn17} & 97.7 & 98.5 & 97.5 & 94.4 & 97.0 \\
OpenVLA-OFT~\citep{kim2025openvlaoft} & 97.6 & 98.4 & 97.9 & 94.5 & 97.1 \\
Fast-WAM~\citep{yuan2026fastwam} & 98.2 & \textbf{100.0} & 97.0 & 95.2 & 97.6 \\
% Being-H0.5~\citep{beingh05} & 99.2 & 98.2 & 99.0 & 96.2 & 98.2 \\
Motus~\citep{bi2025motus} & 96.8 & 99.8 & 96.6 & 97.6 & 97.7 \\
VLA-Adapter~\citep{wang2026vlaadapter} & 99.6 & 99.6 & 98.2 & 96.4 & 98.5 \\
% Qwen-VLA-Instruct & -- & -- & -- & -- & 97.9 \\
% EO1~\citep{qu2025eo1} & \textbf{99.7} & 99.8 & \textbf{99.2} & 94.8 & 98.4 \\
X-VLA~\citep{zheng2025xvla} & 98.2 & 98.6 & 97.8 & 97.6 & 98.1 \\
Cosmos-Policy~\citep{kim2026cosmospolicy} & 98.1 & \textbf{100.0} & 98.2 & 97.6 & 98.5 \\
LingBot-VA~\citep{li2026lingbotva} & 98.5 & 99.6 & 97.2 & \textbf{98.5} & 98.5 \\
Spatial Forcing~\citep{li2025spatialforcing} & 99.4 & 99.6 & 98.8 & 96.0 & 98.5 \\
Xiaomi-Robotics-0~\citep{cai2026xiaomirobotics0} & 98.8 & \textbf{100.0} & 98.8 & 97.2 & 98.7 \\
\midrule
\textbf{UniWAM (Ours)} & \textbf{99.6} & 99.6 & \textbf{99.2} & 98.4 & \textbf{99.2} \\
\bottomrule
\end{tabular}%
}
\end{table}

%% file: table/robotwin_summary.tex
% Standalone benchmark table.
\begin{table}[t]
\centering
\caption{Mean task success rate (\%) on the 50-task RoboTwin~2.0 evaluation.
C2C and C2R denote Clean2Clean and Clean2Random, respectively; gray shading indicates C2R.
Bold and underlined entries denote the highest and second-highest values in each column across both groups.
Methods are grouped into vision-language-action (VLA) and world action model (WAM) families.
}
\label{tab:robotwin-summary}
\small
\setlength{\tabcolsep}{9pt}
\resizebox{0.8\linewidth}{!}{%
\begin{tabular}{lc>{\columncolor{gray!15}}c|c}
\toprule
Method & C2C & C2R & Average \\
\midrule
\rowcolor{gray!10}
\multicolumn{4}{c}{\textbf{VLA}} \\
\midrule
GR00T-N1.7~\citep{nvidia2026gr00tn17}
& 43.6 & 20.7 & 32.2 \\
StarVLA~\citep{community2026starvla}
& 58.1 & 10.6 & 34.4 \\
Xiaomi Robotics-0~\citep{cai2026xiaomirobotics0} & 62.90 & 18.20 & 40.55 \\
Abot-M0~\citep{yang2026abotm0} & 57.40 & 30.36 & 43.88 \\
X-VLA~\citep{zheng2025xvla} & 68.00 & 20.90 & 44.45 \\
Spatial Forcing~\citep{li2025spatialforcing} & 77.20 & 26.74 & 51.97 \\
$\pi_{0.5}$~\citep{physicalintelligence2025pi05} & 70.70 & 46.00 & 58.35 \\
GigaBrain-0.7~\citep{gigabrain2026seven} & 66.80 & \underline{67.90} & 67.35 \\
\midrule
\rowcolor{gray!10}
\multicolumn{4}{c}{\textbf{WAM}} \\
\midrule
AHA-WAM~\citep{cai2026ahawam}
& 64.3 & 3.2 & 33.8 \\
FastWAM~\citep{yuan2026fastwam} & 70.20 & 1.20 & 37.70 \\
X-WAM~\citep{guo2026xwam} & 70.00 & 25.80 & 47.90 \\
4D-WAM~\citep{yang2026four_d_wam}
& \underline{81.5} & 41.8 & 61.7 \\
OpenWAM-$\alpha$~\citep{wang2026openwam}
& \textbf{89.4} & 48.7 & \underline{69.0} \\
\midrule
\textbf{\method~ (ours)} & 75.14 & \textbf{68.32} & \textbf{71.73} \\

% Preserved inactive rows and source notes:
% FastWAM~\citep{yuan2026fastwam} & \underline{77.80} & 1.90 & 39.85 \\
% EventVLA~\citep{yang2026eventvla} & 65.60 & 15.70 & 40.65 \\
% 4D-WAM~\citep{yang2026four_d_wam} & \textbf{81.50} & 41.80 & 61.65 \\
% Additional reported results from arXiv:2609.31394v1, Table 10.

\bottomrule
\end{tabular}%
}
\end{table}

%% file: table/libero_plus_results.tex
\begin{table*}[t]
\centering
\caption{
\textbf{Success rate (\%) under all seven LIBERO-Plus perturbation dimensions (OOD evaluation).}
Bold entries denote the highest value in each column among the listed methods.
Underlined entries denote the second-highest value in each column.
}
\label{tab:libero_plus}
\resizebox{\textwidth}{!}{
\begin{tabular}{lccccccc|c}
\toprule
Method & Camera & Robot & Language & Light & Background & Noise & Layout & Total \\
\midrule

$\pi_0$~\citep{black2024pi0}
& 79.6 & 21.1 & 72.5 & 84.7 & 86.2 & 68.3 & 69.4 & 67.4 \\

GR00T-N1.6~\citep{nvidia2025grootn16}
& 92.6 & 33.5 & 80.1 & 93.6 & 95.4 & 93.6 & 75.0 & 79.4 \\

OpenVLA-OFT~\citep{kim2025openvlaoft}
& 92.8 & 30.3 & 85.8 & 94.9 & 93.9 & 89.3 & 77.6 & 79.5 \\

Spatial Forcing~\citep{li2025spatialforcing}
& 95.2 & 47.9 & 73.5 & 91.2 & 95.6 & 92.2 & 74.8 & 80.5 \\

% VLA-Adapter~\citep{wang2026vlaadapter}
% & \textbf{96.7} & 42.1 & 71.1 & 96.5 & \underline{97.5} & \textbf{96.9} & 74.4 & 81.0 \\

MemoryVLA~\citep{shi2026memoryvla}
& 91.4 & 48.6 & 79.4 & 95.2 & 95.3 & 94.0 & 75.7 & 81.9 \\

% MemoryVLA++~\citep{shi2026memoryvlapp} & \textbf{96.8} & 49.7 & 71.0 & 96.6 & 97.0 & 96.0 & 78.6 & 82.7 \\

$\pi_{0.5}$~\citep{physicalintelligence2025pi05}
& 87.6 & \underline{78.4} & 80.0 & 92.6 & 91.6 & 91.4 & 81.8 & 86.2 \\

ACoT-VLA~\citep{zhong2026acotvla}
& \textbf{96.6} & 70.4 & 79.7 & 95.1 & \underline{97.1} & \underline{95.9} & \textbf{85.0} & 88.0 \\

Kairos~\citep{team2026kairos}
& \underline{95.5} & 72.6 & \underline{86.8} & \underline{97.7} & 95.8 & \textbf{96.8} & 81.5 & \underline{89.0} \\

\midrule
\textbf{UniWAM (Ours)}
& 92.0
& \textbf{89.5}
& \textbf{92.2}
& \textbf{97.9}
& \textbf{97.6}
& 94.3
& \underline{84.6}
& \textbf{92.6} \\

\bottomrule
\end{tabular}
}
\end{table*}

%% file: sec/06_conclusion.tex
\section{Conclusion}
\label{sec:conclusion}
We presented UniWAM, a unified framework that learns physical semantic reasoning, world modeling, and action prediction together. 
To ensure the quality of the training data, we developed a rigorous data cleaning and annotation pipeline for both human egocentric data and robot data. 
Through physical language supervision and component-specific co-training on VQA, human egocentric, and robot data, UniWAM acquires embodied knowledge while preserving pretrained semantic capabilities. 
Post-training further enhances manipulation robustness by reducing reliance on visual predictions, while substantially accelerating inference without compromising performance.
Extensive evaluations demonstrate strong task performance, robustness, generalization, instruction following, and long-horizon execution. 
The uncovered log-linear scaling law of human-robot co-training suggests that pretraining a unified world-action architecture on heterogeneous data offers a promising path toward generalist robot models.

%% file: sec/appendix.tex
\section{Overview}
In this supplementary material, we provide the following information:
\begin{itemize}
    \item Section~\ref{app:real-world} describes the real-world tasks and experimental setup.
    \item Section~\ref{app:action-generation-steps} compares action-generation strategies across inference-step counts.
    \item Section~\ref{app:physical-action-templates} describes the physical action template shared across robot and human data.
    % \item Sections~\ref{app:human-data} and~\ref{app:egoant} describe the human egocentric datasets and our automatic annotation pipeline.
    \item Section~\ref{app:robotwin-evaluation} details the RoboTwin 2.0 evaluation protocol.
    \item Section~\ref{app:task-results} reports task-level RoboTwin 2.0 results.
    \item Section~\ref{app:config} summarizes implementation details.
    % \item Section~\ref{app:data-process} describes robot-data processing and quality control.
\end{itemize}

\section{Real-world Tasks}
\label{app:real-world}
\begin{figure}[t]
    \centering
    \includegraphics[width=0.6\textwidth]{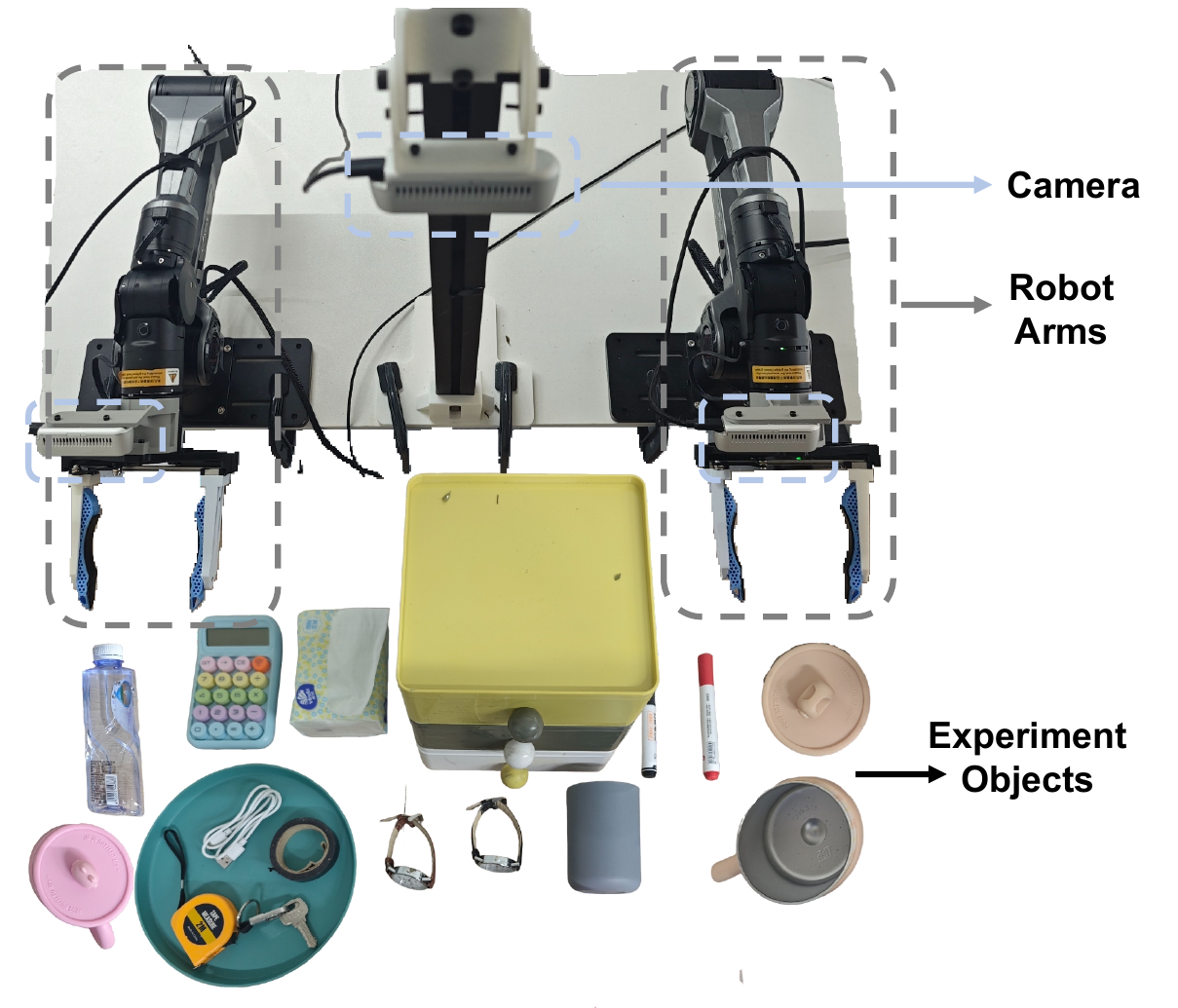}
    \caption{\textbf{Real-world experimental setup.}
Our setup consists of two robot arms, three cameras for observation, and a diverse set of everyday objects used in the manipulation experiments.
    }
    \label{setting_real_task}
\end{figure}

% \begin{figure}[ht]
%     \centering
%     \includegraphics[width=\textwidth]{figure/real_task.pdf}
%     \caption{
%     \textbf{Overview of our real-world evaluation tasks.}
% The top panel shows representative execution sequences for four instruction-following tasks:
% Pick-Anything, Diverse-Interaction, Place-Relative, and Drawer Storage.
% The bottom panel presents a long-horizon task,
% where the robot sequentially completes multiple manipulation stages.}
% \label{overview_real_task}
% \end{figure}

%附录:实验物体如图,包含:计算器, 抽纸, 抽屉, 不同颜色的马克笔,盘子, 卷尺, 钥匙, 数据线, 胶带, 橙色杯子, 粉色杯子, 杯盖, 棕色手表, 黑色手表, 漱口杯, 矿泉水瓶.所有任务均是双臂任务,目标物体由离得近的那条机械臂进行抓取. 每个任务我们评估40次.待插入图片
\noindent\textbf{Objects and Robot Setup.}
Our real-world experiments involve a diverse set of household objects, as illustrated in Fig.~\ref{setting_real_task}, including a calculator, tissue box, drawer, markers of different colors, plate, tape measure, keys, data cable, tape, orange cup, pink cup, cup lid, brown watch, black watch, plastic cup, and water bottle. All experiments are conducted using the dual-arm setup described in the main text. For object-centric manipulation, the arm spatially closer to the target object is responsible for grasping it, while both arms can participate when bimanual coordination is required. Each task is evaluated over 40 trials.
%Pick-Anything任务场景任选八个物体随机散落在桌面上和一个固定的盘子,指令是抓取xxx到盘子里.

\paragraph{Instruction-following tasks.}

\noindent\textbf{Pick-Anything.}
Each scene contains eight objects randomly selected from the object set and scattered across the tabletop, together with a plate placed at a fixed location. The instruction follows the form ``pick up [target object] and place it on the plate.'' This task primarily evaluates object-level language grounding and the ability to distinguish the instructed target from multiple distractors.
%Diverse-Interaction任务场景有杯子,杯盖,计算器和两个其他可抓取的物体和盘子.指令是抓xx物体到盘子里或给杯子盖杯盖或是启动计算器.这里涉及到同一物体(杯子或计算器)的不同操作,需要模型真正理解指令.

\noindent\textbf{Diverse-Interaction.}
Each scene contains a cup, a cup lid, a calculator, two additional graspable objects, and a plate. The instructions require the robot to perform different interactions according to the specified object and action, including placing a target object on the plate, placing the lid onto the cup, or activating the calculator. The same object can be associated with different manipulation behaviors across instructions. The task therefore requires the policy to jointly ground both object semantics and action semantics, rather than relying solely on object recognition.

%Place-Relative任务场景有任意五个物体在桌面上, 指令是抓取xxx物体放到另一xxx物体的左边或右边.
\noindent\textbf{Place-Relative.}
Each scene contains five objects placed at varying locations on the tabletop. The robot is instructed to pick up one specified object and place it either to the left or to the right of another specified object, e.g., ``place the pink cup to the left of the calculator.'' Successful execution requires simultaneously identifying both the source and reference objects and grounding the specified spatial relation. This task evaluates compositional object grounding and spatial-relation understanding.
%Drawer Storage任务场景有一个二层抽屉和至多三个物体,指令是抓取xxx物体到第一或第二层抽屉.

\noindent\textbf{Drawer Storage.}
The scene contains a two-level drawer and up to three candidate objects. The instruction specifies both a target object and a target drawer level, requiring the robot to place the designated object into either the upper or lower drawer. Completing the task involves object identification, drawer-level grounding, sequential manipulation, and bimanual coordination, as one arm interacts with the drawer while the other handles the target object.

% \paragraph{Long-horizon Tasks.}
% The long-horizon task is composed of 6 subtasks: (1) placing the first cup onto a plate, (2) placing the second cup onto the plate, (3) opening the drawer, (4) placing the designated daily-use object or tool inside its corresponding drawer, (5) closing the drawer, and (6) placing a marker into its designated holder.

\section{Action Generation with Different Inference Steps}
\label{app:action-generation-steps}
\begin{figure}[htbp]
    \centering
    \includegraphics[width=\linewidth]{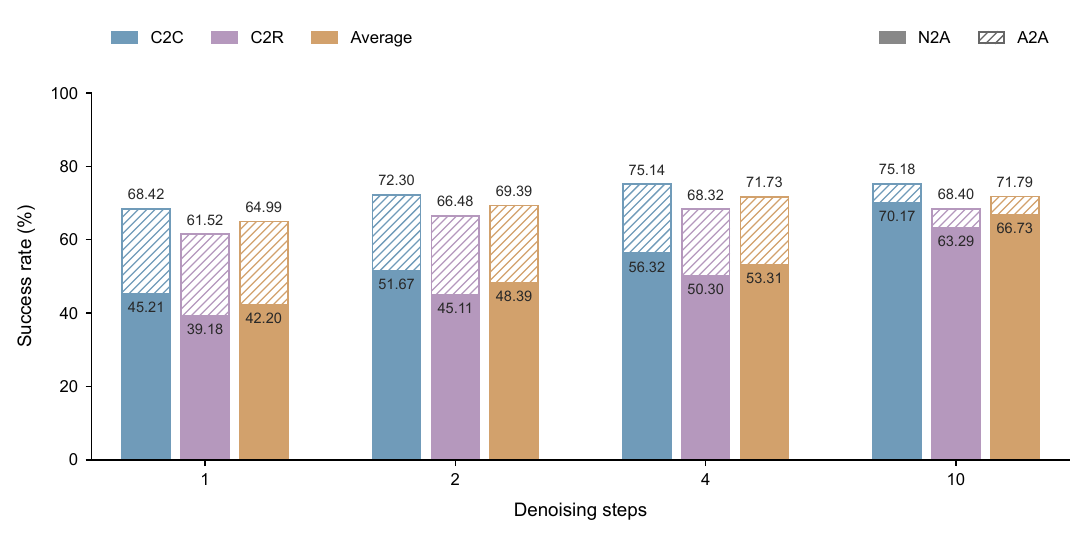}
    \caption{\textbf{Action generation performance across denoising steps.}
    N2A and A2A denote noise-to-action and action-to-action generation, respectively..
    The endpoint of the solid portion corresponds to the N2A success rate, while the endpoint of the hatched portion corresponds to the A2A success rate.
    All success rates are reported in percent.}
    \label{fig:action-generation-steps}
\end{figure}
The action denoising process is illustrated in \Cref{fig:a2a}.
As shown in \Cref{fig:action-generation-steps}, A2A achieves strong performance with only 2--4 inference steps, even outperforming N2A with ten steps. This demonstrates that A2A can achieve high action-generation quality with fewer denoising steps. We therefore use four-step A2A as the default inference setting.

\section{Physical Action Templates}
\label{app:physical-action-templates}
We use a shared physical-language template to describe end-effector action
changes in robot and human data. At each timestep, the template summarizes
the net changes over a sliding action window; absolute pose inputs are first
converted to per-step changes. Translation magnitudes are expressed in
centimeters and rounded to the nearest integer, while rotation magnitudes are
expressed in degrees and rounded to the nearest $10$ degrees. Positive and
negative changes map to ``move forward'' and ``move back'' along $x$,
``move up'' and ``move down'' along $z$, and ``move left'' and ``move right''
along $y$. Rotations are described as tilting left/right (roll), tilting
back/forward (pitch), or rotating counterclockwise/clockwise (yaw). Components
with no change are omitted. When gripper values are available, the final value
in the window determines whether the text says ``open gripper'' (value at
least $0.5$) or ``close gripper'' (value below $0.5$). For bimanual actions,
the summaries are prefixed with ``Left arm:'' and ``Right arm:''.
This shared template provides a consistent language interface for action
supervision across the two data sources.

The following are illustrative template outputs, not examples taken from
recorded demonstrations:
\begin{quote}
\emph{Single arm:} ``move forward 3 cm, move left 2 cm, tilt left 10 degrees,
rotate counterclockwise 20 degrees, open gripper.''\\
\emph{Bimanual:} ``Left arm: move up 4 cm, close gripper. Right arm: move back
2 cm, open gripper.''
\end{quote}

\input{table/robotwin_task_results}

\section{Details of Evaluation on RoboTwin 2.0}
\label{app:robotwin-evaluation}
RoboTwin 2.0 is a benchmark consisting of 50 bimanual manipulation tasks~\citep{chen2025robotwin2}. 
We train the model on all 50 tasks with 50 clean demonstrations per task and evaluate it on clean (C2C) and randomized (C2R) settings individually for 100 episodes, where the latter evaluates the generalization abilities.

\section{RoboTwin 2.0 Task-Level Results}
\label{app:task-results}

Table~\ref{tab:robotwin-task-results} reproduces every task-level value supplied
in the experiment record. The table is the authoritative location for these
individual percentages; the main text reports only the values needed to expose
aggregate performance and important failure boundaries.

\section{Implementation Details}
\label{app:config}
\paragraph{Model architecture.}
\method~combines Qwen3-VL-2B-Instruct as the physical reasoner and
Wan2.2-TI2V-5B as the world generator with an action predictor in a
Mixture-of-Transformers architecture. The three experts comprise 30 layers
each. The world generator has a hidden dimension of 3,072 and 24 attention
heads; the physical reasoner and action predictor have hidden dimensions of
512 and 1,024, respectively. The model uses 14-dimensional robot states and
actions, predicts eight future frames at $384\times320$ resolution, and
generates action chunks of length 16.

\paragraph{Pretraining.}
Pretraining runs for seven days on 32 NVIDIA H800 GPUs. The joint objective
weights the video, action, and language losses in the ratio $1:1:0.1$:
\begin{equation}
  \mathcal{L}_{\mathrm{pretrain}}
  =\mathcal{L}_{\mathrm{video}}
   +\mathcal{L}_{\mathrm{action}}
   +0.1\,\mathcal{L}_{\mathrm{language}}.
\end{equation}

\paragraph{Post-training.}
For RoboTwin post-training, we use a batch size of 40 for 200,000 optimization
steps. We optimize with AdamW, a learning rate of $5\times10^{-5}$, and weight
decay of 0.01. The learning-rate schedule is linear with 200 warmup steps;
gradient norm is clipped at 0.5. The Wan and VLM backbones use bfloat16
precision. Future-frame noise augmentation is applied with probability 0.5,
with the visual latent retention scale sampled uniformly from $[0.5,1.0]$.
The RoboTwin run uses 8 NVIDIA H800 GPUs for 12 hours; the LIBERO run uses
8 NVIDIA H800 GPUs for 20 hours.

\paragraph{Inference.}
The action flow is initialized from a recent action-history chunk of length
16 with Gaussian perturbation of standard deviation 0.02, while the
future video latents are initialized from Gaussian noise. We use 2--4 flow
integration steps at inference.

%% file: table/robotwin_task_results.tex
\begin{table}[t]
\centering
\small
\caption{Task-level UniWAM success rates (\%) on RoboTwin~2.0. C2R denotes
clean2rand and C2C denotes clean2clean. C2C and C2R values are reproduced from the supplied
experiment record. Average is their arithmetic mean.}
\label{tab:robotwin-task-results}
\resizebox{\textwidth}{!}{%
\begin{tabular}{lrrr@{\hspace{1.4em}}lrrr}
\toprule
Task & C2R & C2C & Average & Task & C2R & C2C & Average \\
\midrule
\texttt{adjust\_bottle} & 72 & 72 & 72.00 & \texttt{place\_can\_basket} & 35 & 36 & 35.50 \\
\texttt{beat\_block\_hammer} & 56 & 65 & 60.50 & \texttt{place\_cans\_plasticbox} & 86 & 97 & 91.50 \\
\texttt{blocks\_ranking\_rgb} & 67 & 64 & 65.50 & \texttt{place\_container\_plate} & 93 & 97 & 95.00 \\
\texttt{blocks\_ranking\_size} & 38 & 36 & 37.00 & \texttt{place\_dual\_shoes} & 77 & 87 & 82.00 \\
\texttt{click\_alarmclock} & 92 & 100 & 96.00 & \texttt{place\_empty\_cup} & 94 & 98 & 96.00 \\
\texttt{click\_bell} & 99 & 100 & 99.50 & \texttt{place\_fan} & 69 & 76 & 72.50 \\
\texttt{dump\_bin\_bigbin} & 94 & 96 & 95.00 & \texttt{place\_mouse\_pad} & 55 & 59 & 57.00 \\
\texttt{grab\_roller} & 100 & 100 & 100.00 & \texttt{place\_object\_basket} & 50 & 54 & 52.00 \\
\texttt{handover\_block} & 18 & 24 & 21.00 & \texttt{place\_object\_scale} & 64 & 69 & 66.50 \\
\texttt{handover\_mic} & 81 & 89 & 85.00 & \texttt{place\_object\_stand} & 78 & 81 & 79.50 \\
\texttt{hanging\_mug} & 16 & 16 & 16.00 & \texttt{place\_phone\_stand} & 64 & 75 & 69.50 \\
\texttt{lift\_pot} & 92 & 100 & 96.00 & \texttt{place\_shoe} & 79 & 79 & 79.00 \\
\texttt{move\_can\_pot} & 46 & 46 & 46.00 & \texttt{press\_stapler} & 96 & 98 & 97.00 \\
\texttt{move\_pillbottle\_pad} & 71 & 78 & 74.50 & \texttt{put\_bottles\_dustbin} & 38 & 44 & 41.00 \\
\texttt{move\_playingcard\_away} & 93 & 90 & 91.50 & \texttt{put\_object\_cabinet} & 28 & 40 & 34.00 \\
\texttt{move\_stapler\_pad} & 53 & 58 & 55.50 & \texttt{rotate\_qrcode} & 41 & 46 & 43.50 \\
\texttt{open\_laptop} & 75 & 90 & 82.50 & \texttt{scan\_object} & 33 & 42 & 37.50 \\
\texttt{open\_microwave} & 44 & 76 & 60.00 & \texttt{shake\_bottle} & 71 & 89 & 80.00 \\
\texttt{pick\_diverse\_bottles} & 78 & 76 & 77.00 & \texttt{shake\_bottle\_horizontally} & 70 & 92 & 81.00 \\
\texttt{pick\_dual\_bottles} & 85 & 92 & 88.50 & \texttt{stack\_blocks\_three} & 52 & 74 & 63.00 \\
\texttt{place\_a2b\_left} & 64 & 65 & 64.50 & \texttt{stack\_blocks\_two} & 91 & 91 & 91.00 \\
\texttt{place\_a2b\_right} & 80 & 85 & 82.50 & \texttt{stack\_bowls\_three} & 82 & 88 & 85.00 \\
\texttt{place\_bread\_basket} & 81 & 84 & 82.50 & \texttt{stack\_bowls\_two} & 91 & 94 & 92.50 \\
\texttt{place\_bread\_skillet} & 68 & 81 & 74.50 & \texttt{stamp\_seal} & 70 & 87 & 78.50 \\
\texttt{place\_burger\_fries} & 76 & 95 & 85.50 & \texttt{turn\_switch} & 70 & 86 & 78.00 \\
\midrule
\multicolumn{1}{l}{Mean} & 68.32& 75.14 & 71.73 &  &  &  &  \\
\bottomrule
\end{tabular}%
}
\end{table}